\documentclass[twoside]{article}

\usepackage[accepted]{aistats2021}
\usepackage[round]{natbib}

\usepackage[T1]{fontenc} 
\usepackage{microtype}
\usepackage{graphicx}
\usepackage{subfigure}
\usepackage{booktabs} 

\usepackage{xcolor}
\usepackage[nice]{nicefrac}

\usepackage{amsmath}
\usepackage{amssymb}
\usepackage{algorithm}
\usepackage{algorithmic}
\usepackage{caption}

\DeclareMathOperator*{\argmin}{argmin}
\DeclareMathOperator*{\argmax}{argmax}
\DeclareMathOperator*{\cov}{cov}

\newcommand{\Xt}{\mathbf{X}_t}
\newcommand{\yt}{\mathbf{y}_t}
\newcommand{\Pt}{\mathbf{\Phi}_t}
\newcommand{\wt}{\mathbf{w}_t}
\newcommand{\N}{\mathrm{Normal}}

\newcommand{\z}{\mathbf{z}}
\newcommand{\I}{\mathbf{I}}
\newcommand{\nul}{\mathbf{0}}
\newcommand{\E}{\mathbb{E}}
\newcommand{\x}{\mathbf{x}}

\newcommand{\An}{\Diag(\mathbf{\alpha}_{\textit{new}})}
\newcommand{\Fn}{\NDn}

\newcommand{\yn}{\mathbf{y}_{\textit{new}}}

\newcommand{\wn}{\mathbf{w}_{\textit{new}}}
\newcommand{\an}{\alpha_{\textit{new},i}}
\newcommand{\bn}{\beta_{\textit{new}}}
\newcommand{\NDn}{\mathbf{\Phi}_{\textit{new}, b\downarrow}}
\newcommand{\n}{\textit{new}}

\DeclareMathOperator{\Diag}{Diag}

\usepackage{hyperref}

\renewcommand{\E}{\mathbb{E}}

\renewcommand{\I}{\mathbf{I}}

\makeatletter
\newcommand{\printfnsymbol}[1]{%
  \textsuperscript{\@fnsymbol{#1}}%
}

\begin{document}

\twocolumn[

\aistatstitle{Hyperparameter Transfer Learning \\with Adaptive Complexity}

\aistatsauthor{ Samuel Horv\'{a}th \And Aaron Klein \And  Peter Richt\'{a}rik \And C\'{e}dric Archambeau}

\aistatsaddress{ KAUST \And  Amazon Web Services  \And KAUST \And Amazon Web Services } ]

\begin{abstract}
Bayesian optimization (BO) is a sample efficient approach to automatically tune the hyperparameters of machine learning models. 
In practice, one frequently has to solve similar hyperparameter tuning problems sequentially. For example, one might have to tune a type of neural network learned across a series of different classification problems. Recent work on multi-task BO exploits knowledge gained from previous tuning tasks to speed up a new tuning task. 
However, previous approaches do not account for the fact that BO is a sequential decision making procedure. Hence, there is in general a mismatch between the number of evaluations collected in the current tuning task compared to the number of evaluations accumulated in all previously completed tasks.
In this work, we enable multi-task BO to compensate for this mismatch, such that the transfer learning procedure is able to handle different data regimes in a principled way. We propose a new multi-task BO method that learns a set of ordered, non-linear basis functions of increasing complexity via nested drop-out and automatic relevance determination. Experiments on a variety of hyperparameter tuning problems show that our method improves the sample efficiency of recently published multi-task BO methods.
\end{abstract}

\section{Introduction}

Bayesian Optimization (BO)~\citep{ review_shahriari2015taking, review_frazier2018tutorial} is a well-established framework to optimize an expensive black-box function $f: \mathcal{X} \rightarrow \mathbb{R} $ with a minimal number of function evaluations:
\begin{equation}
    \label{eq:bo}
    \x^{\star} = \argmin_{\x \in \mathcal{X}} f(\x),
\end{equation}
where $\mathcal{X} \subset \mathbb{R}^D$ denotes the configuration space. We assume $f(\x)$ is only observed through noisy evaluations $y \sim \N(f(\x), \sigma)$. Intuitively, BO searches for the global optimum $\x^{\star}$ by evaluating a sequence of promising candidates $\x_1, \ldots, \x_N$. At each step, candidates are scored according to an exploration-exploitation criterion based on the evaluations collected so far. In the case of hyperparameter optimization (HPO), $f$ denotes the validation error and $\x$ is a vector of hyperparameters over which to optimize.

Let $\mathcal{D}= \{(\x_{n}, y_{n})\}_{n=1}^{N}$ be a set of evaluated hyperparameter configurations and their associated validation error. BO uses a probabilistic model $p(f \mid \mathcal{D})$ of the unknown objective function $f$ to select the next configuration by optimizing an acquisition function $a: \mathcal{X} \rightarrow \mathbb{R}$ that performs the explore-exploit trade-off. 
The acquisition function only relies on the model $p(f \mid \mathcal{D})$ and, hence, is cheap-to-evaluate compared to the objective function.
Typical choices for the probabilistic model are Gaussian Processes (GPs)~\citep{gp_jones2001taxonomy, gp_snoek}, random forests~\citep{random_forrest}, tree parzen estimators~\citep{tpe_bergstra2011algorithms}, and Bayesian neural networks~\citep{dngo_snoek2015scalable, springenberg2016bayesian,ablr_perrone2018scalable}. Arguably, the most common acquisition function is the expected improvement (EI) \citep{BO_mockus1978application}. EI selects the next candidate to evaluate as the one with the highest expected improvement $a(\x) = \E[ \max\{0, f^* - f(\x)\} ]$, where $f^*$ is the best evaluation observed so far. Numerous other acquisition functions have been proposed \citep{aq_kushner1964new, aq_frazier2009knowledge, aq_srinivas2009gaussian, aq_hoffman2011portfolio, aq_hennig2012entropy}. 

\begin{figure*}[t]
  \subfigure[\texttt{ABLR}]{
  \includegraphics[width=0.24\textwidth]{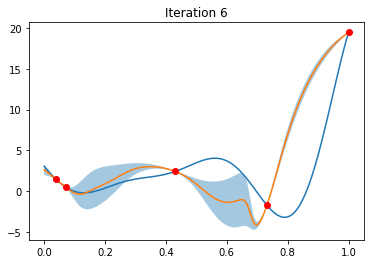}
  \includegraphics[width=0.24\textwidth]{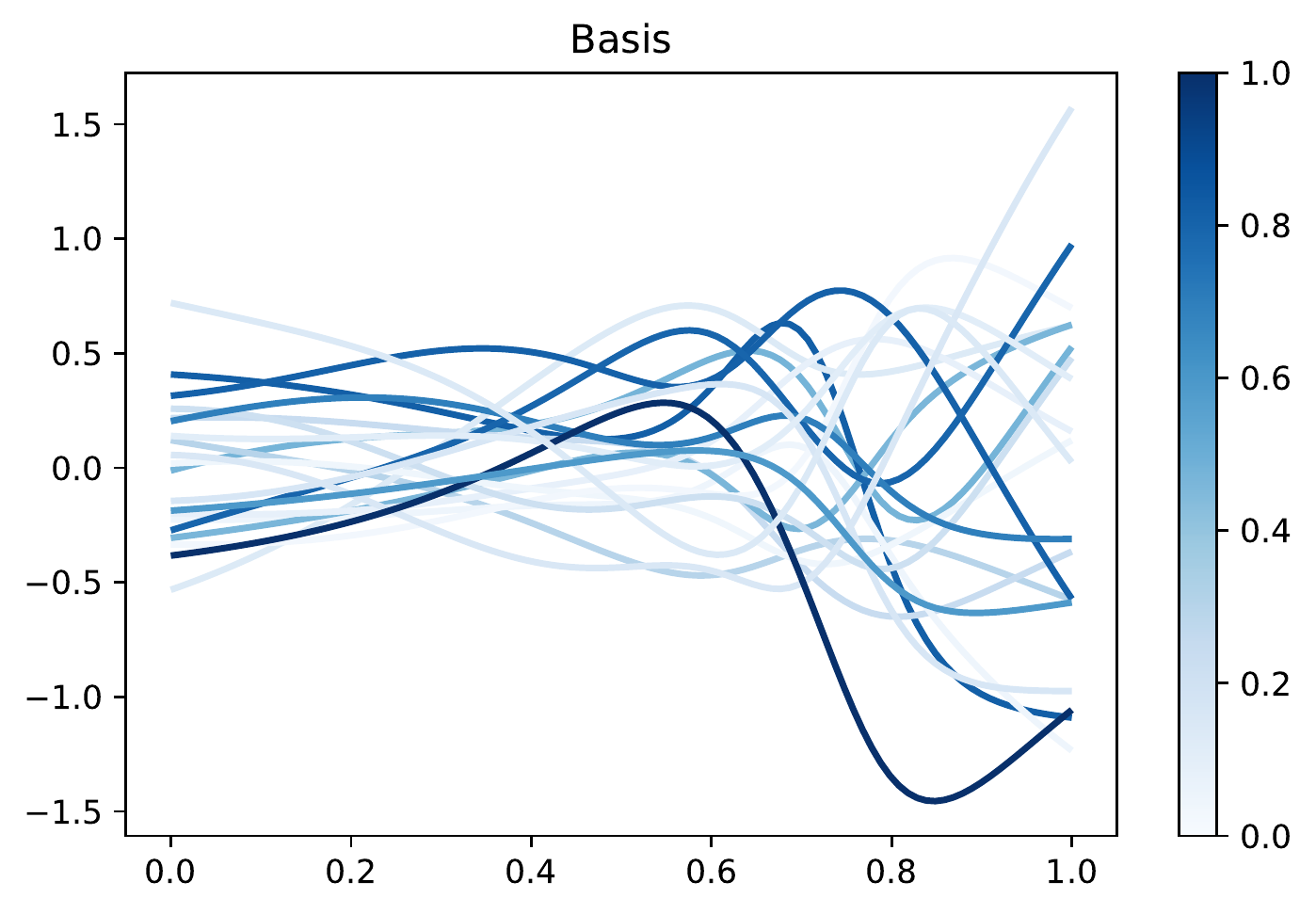}}
\hfill
  \centering
  \subfigure[\texttt{ABRAC}]{
  \includegraphics[width=0.24\textwidth]{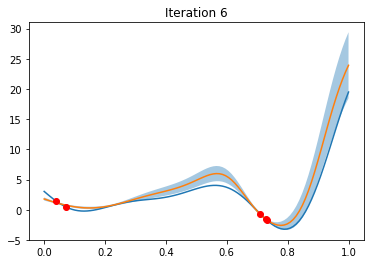}
  \includegraphics[width=0.24\textwidth]{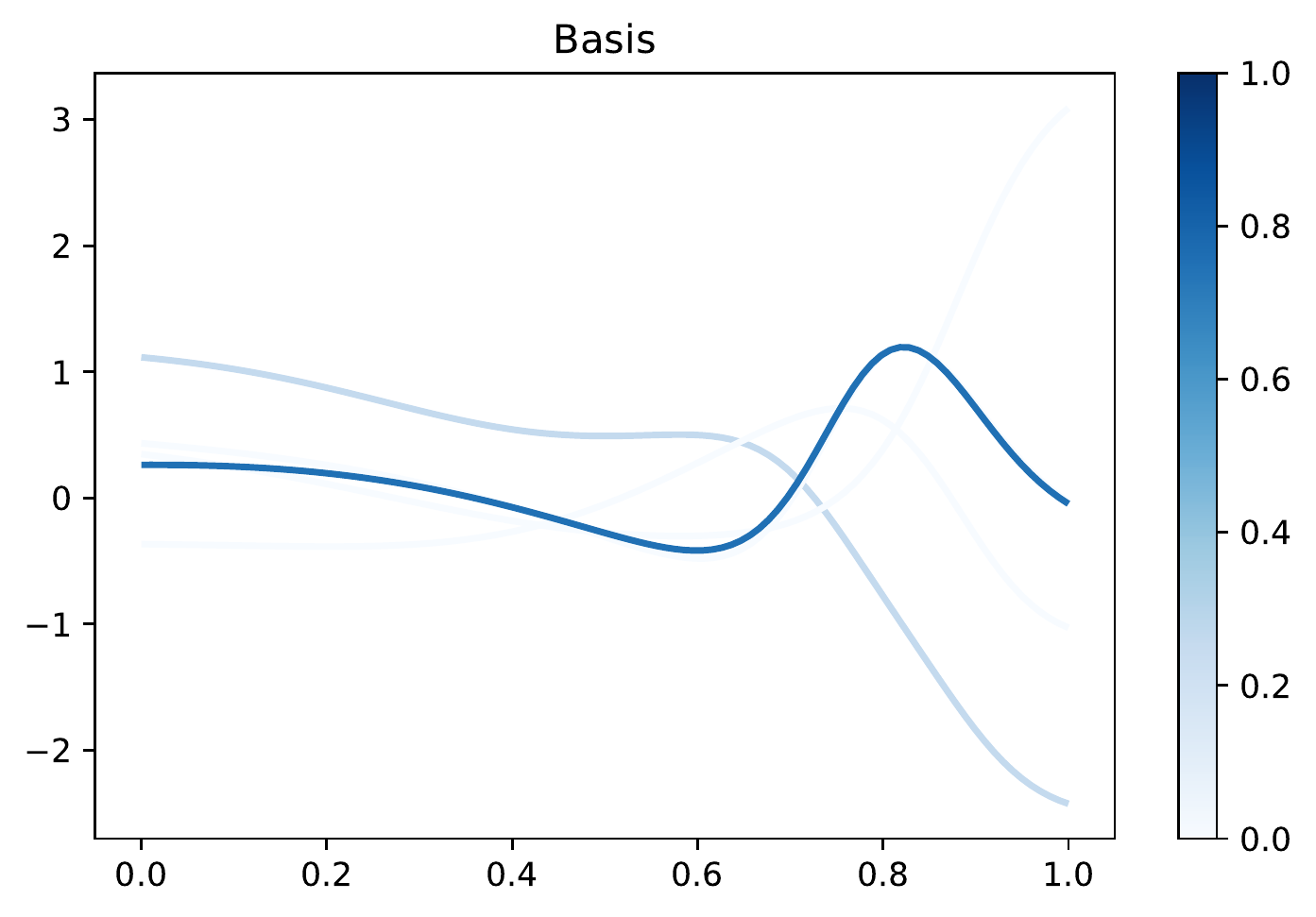}}
\hfill
  \caption{ Comparison of the surrogate objectives along with their basis functions obtained by (a) conventional multi-task BO, implimented by ABLR \citet{ablr_perrone2018scalable}, and (b) multi-task BO with adaptive complexity, ABRAC (see Section~\ref{sec:abrac}), which restricts the number of basis functions that can be activated at transfer. In the first and the third panel the target function is shown in blue, the five evaluations are the red points (which includes the same $3$ initial random points for both methods); the surrogate predictive mean is shown in orange with two standard deviations in transparent light blue. The second and fourth panel show the basis functions with their color representing their relative weight used at iteration 6 of the BO algorithm. It can be observed that overfitting occurs in (a) due to the large number of basis functions used, while no overfitting occurs in (b) by restricting that number (note that there are more than 2 basis functions (5) for (b), but they are almost invisible as their relative weights are very close to zero).}
  \label{fig:1d_unc_intro}
\end{figure*}

In this work, we focus on multi-task BO. We assume that we have already completed $T$ related HPO tasks $\{f_1, \dots f_T\}$ that share the same configuration space $\mathcal{X}$. Our goal is to leverage knowledge acquired in previous HPO tasks to accelerate the optimization of a new HPO task $f_{T+1}$. 
In multi-task learning \citep[see, e.g.,][]{argyriou2006nips} the number of training data available for each task is typically of the same order. Multi-task BO, similar to BO, differs from conventional multi-task learning as it is an instantiation of Bayesian sequential decision making. This means that the training data are collected \emph{sequentially} during the training procedure, which in turns means that the number of data associated to the new task is initially much smaller than the number data that were collected in each previous task. Not accounting for this in practice is problematic as the transfer learning procedure is not adapted to the amount of information available in the target task as illustrated in Figure~\ref{fig:1d_unc_intro}.

Our proposed method achieves transfer learning by learning a set of shared basis functions across multiple tasks. Following \citet{ablr_perrone2018scalable}, we use an underlying shared dense neural network to learn the basis functions on top of which we place task-specific Bayesian linear regression head models. 

We introduce a novel regularization strategy based on nested dropout~\citep{nested_rippel2014learning}, which we apply to the final layer of the neural network. This enforces the network to learn an ordered set of basis functions of increasing complexity, which in turn allows us to model different data regimes. We further leverage automatic relevance determination (ARD) to automatically determine the number of basis functions to be activated for transfer learning~\citep{mackay}.

Next, we summarize our contributions and relate our work to the recent literature on multi-task BO. In Section~\ref{sec:ablr} and \ref{sec:nested} we review, respectively, multi-head ABLR for multi-task BO and nested dropout. In Section~\ref{sec:abrac}, we describe our new multi-task BO method, which is able to learn ordered features for a multi-head ABLR model. We present an empirical evaluation in Section~\ref{sec:experiments}, showing that the sample efficiency of our proposed approach is superior to conventional multi-task BO. Finally, we provide a conclusion and future directions in Section~\ref{sec:disc}.

\begin{figure*}[!t]
\centering
\includegraphics[width=0.39\textwidth]{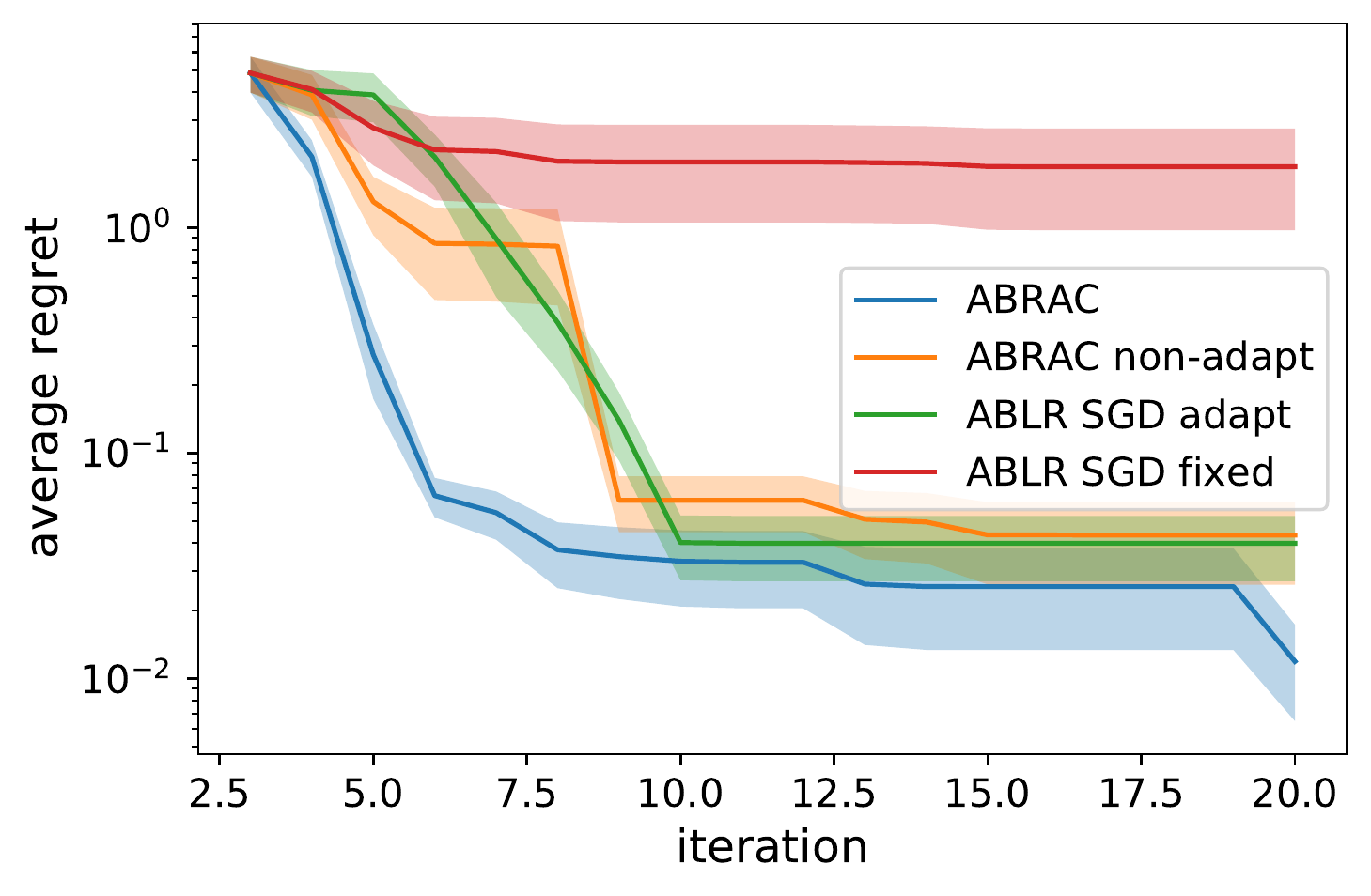}
\includegraphics[width=0.39\textwidth]{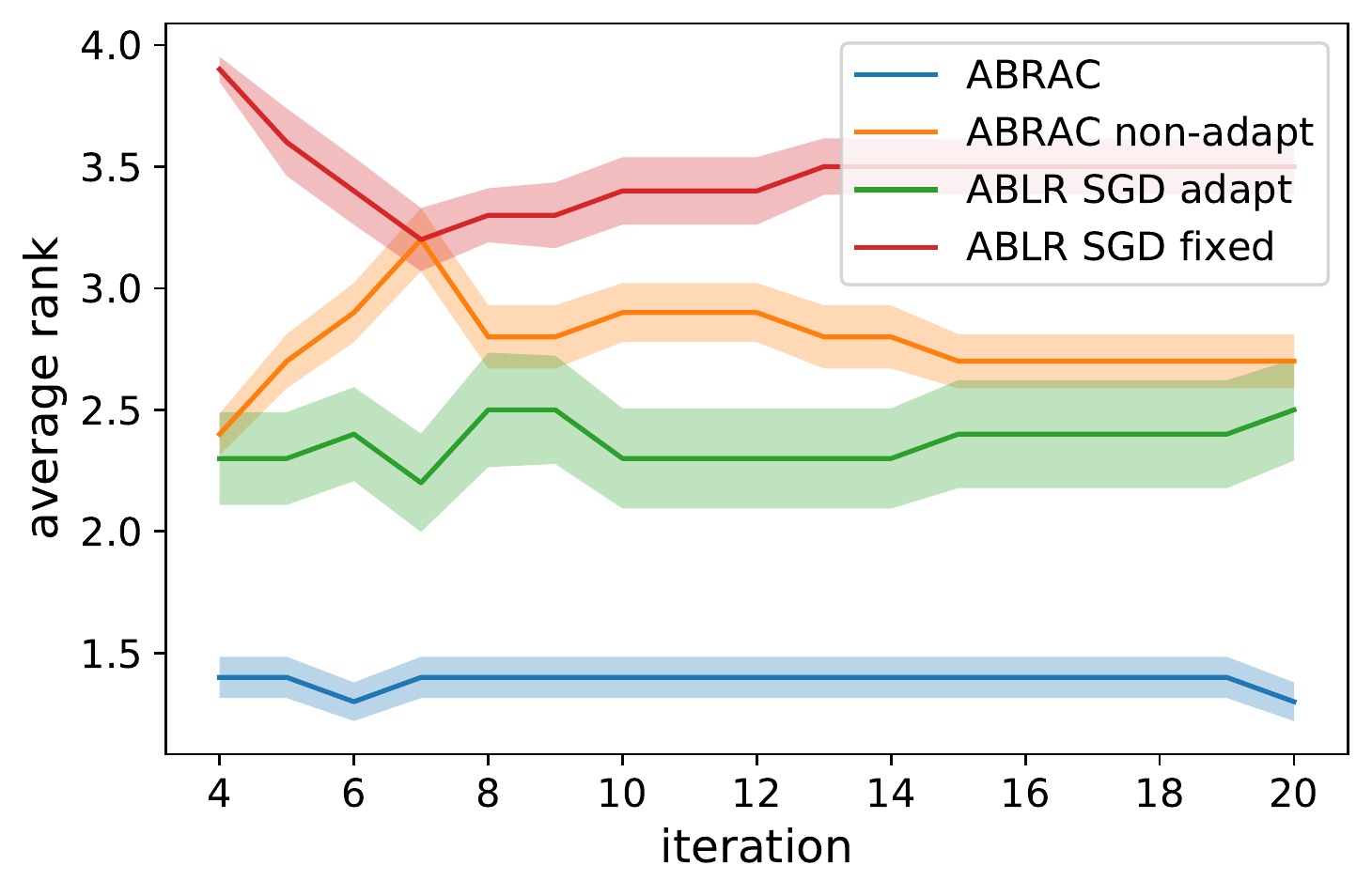}
\caption{Comparison of the average regret and rank on Forrester functions for 5 different methods. Solid lines display average over all the runs of given method, transparent areas depict $2$ standard deviations.}
\label{fig:ablation}
\end{figure*}

\textbf{Contributions.}
Our contributions can be summarized as follows:
\begin{itemize}
\item We use nested dropout to learn an ordered set of features for transfer learning in the context of BO. To the best of our knowledge, no other multi-task learning approach proposed in the literature is able to learn features that take the adaptive complexity of the target task into account. 
\item We use ARD to automatically determine which basis functions to activate at transfer in a data-driven fashion. Hence, the resulting transfer learning model is able to adapt its capacity to the amount of data available in the target task.\\
\item We show that we can improve the sample efficiency of multi-task BO and avoid overfitting in low data regimes without hurting the transfer learning performance in high data regimes.
\end{itemize}

Our multi-task BO method scales linearly with respect to the number of data points collected in the target BO task. This is achieved by adopting a two-step procedure: (1) a more expensive offline step consists in extracting knowledge in an ordered fashion from the completed BO tasks; (2) a fast online step exploits this knowledge to solve the new BO task data efficiently. The complexity of (1) is linear in the evaluations collected across all BO tasks, while the complexity (2) is linear in the data collected in the target BO task only. 

\textbf{Related Work.}
\citet{gp_feurer2015initializing} defined a set of hand-designed meta-features to describe similarity between tasks. 
Initial configurations to warm-start Bayesian optimization are then selected based on the euclidean distance of the meta-features of the current task to previous tasks.
Instead of manually defined features to describe tasks, \citet{law-nips19} proposed a feature extractor based on neural networks that operates directly on the underlying dataset.

Instead of learning features explicitly, \citet{swersky2013multi} proposed a discrete kernel to enable GP-based BO to model the correlation across tasks. While this model does not make any strong assumptions, it does not scale well with the number of tasks.
In a similar vein, \citet{springenberg2016bayesian} proposed Bayesian neural networks for BO and showed that, by using a linear embedding that is learned together with the other parameters of the neural network, one can model the objective function across tasks.
Instead of an additional input to the model, \citet{ablr_perrone2018scalable} considered a multi-head neural network architecture with one head per tasks. To obtain uncertainty estimates, they followed \citet{dngo_snoek2015scalable} by using Bayesian linear regression head models. \citet{salinas-icml20} used a semi-parametric Gaussian Copula distribution to map hyperparameter configurations to quantiles of the objective function across different tasks.
Most recently, \citet{moss-arxiv20} proposed a new information-theoretic acquisition function for Multi-Task Bayesian optimization which is much faster to compute than previous existing approaches.
Compared to our method, none of these transfer learning approaches learn an ordered set of features of variable complexity.

Also related to our approach is the idea to learn low-dimensional embeddings for high dimensional optimization tasks~\citep{moriconi-arxiv19,nayebi-icml19}. However, the key difference is that these embeddings are only learned on a single task without any transfer across tasks.

There also has been a large body of work on multi-task feature learning since the seminal work of \citet{argyriou2006nips}. Most of the published work focusses on learning a sparse set of features and methods that avoid negative transfer in multi-task learning~\citep{guo2014nips}. More recently, there has been a growing interest in transfer learning in the context of deep learning~\citep{ruder2017arxiv}. To the best of our knowledge, none of these techniques attempt to learn an ordered set of multi-task features.

\section{Multi-head ABLR for Multi-task Bayesian Optimization}\label{sec:ablr}

\begin{figure*}[t]
\centering
\includegraphics[width=0.39\textwidth]{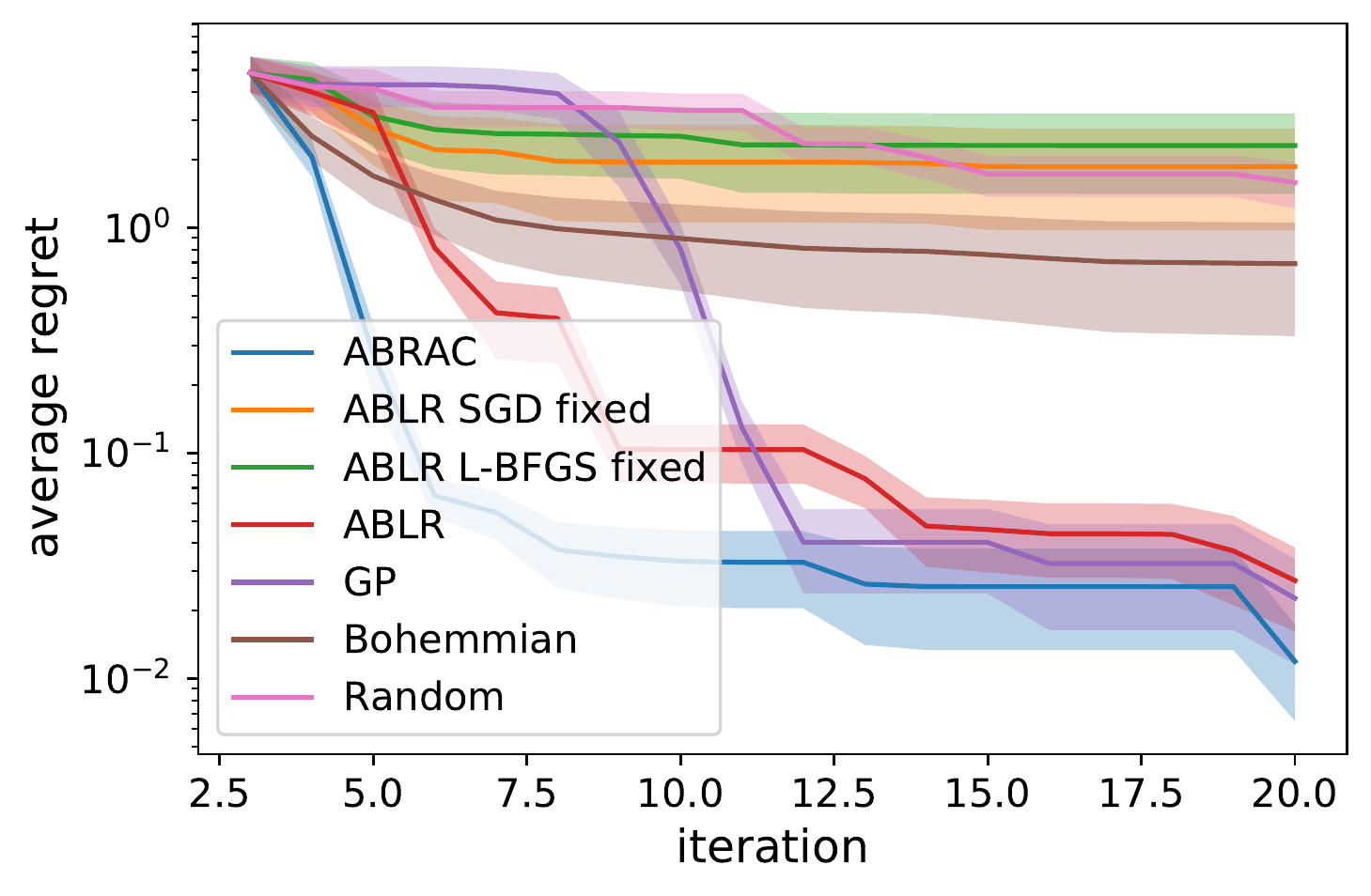}
\includegraphics[width=0.39\textwidth]{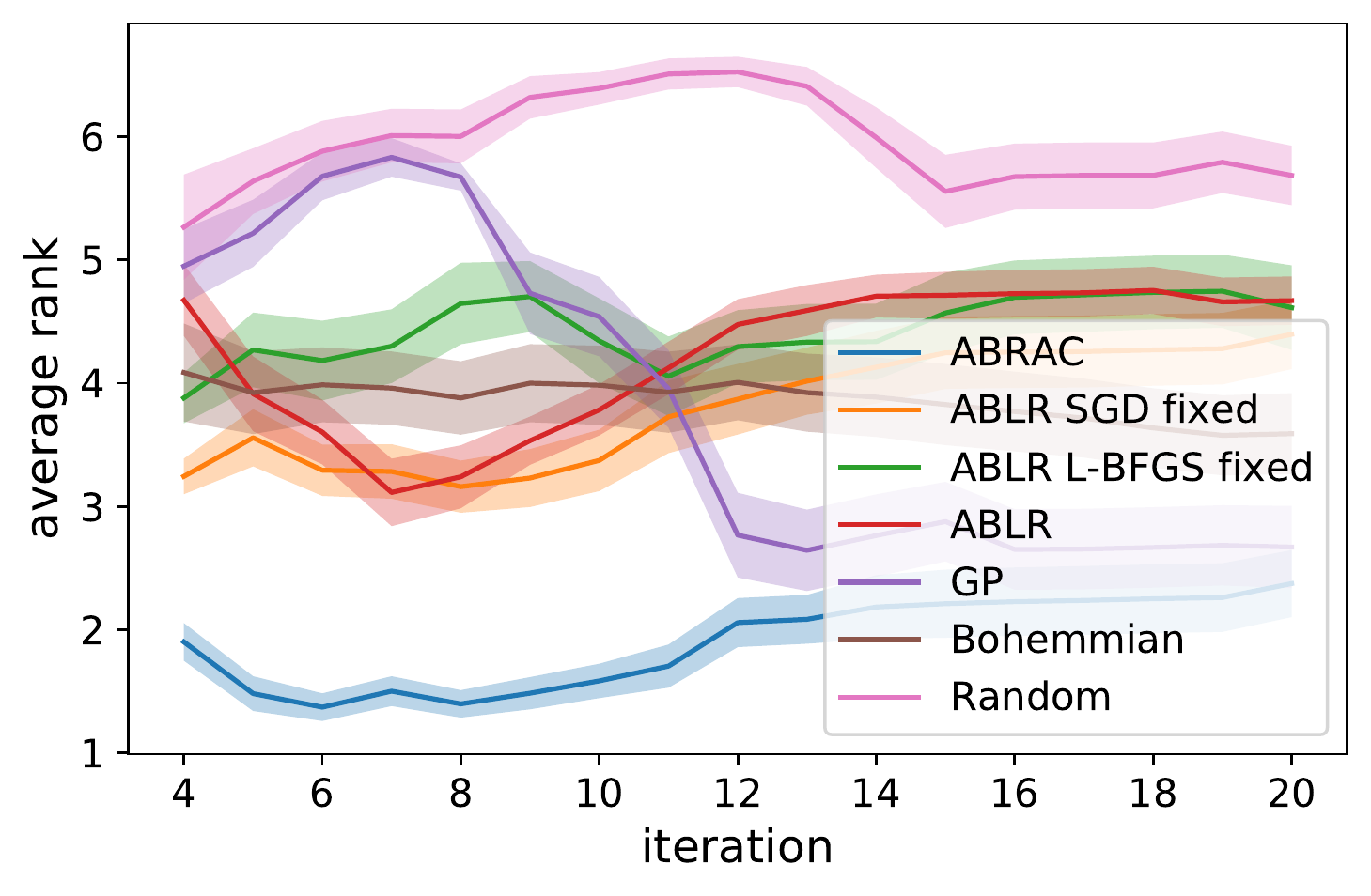}
\caption{Comparison of the average regret and rank on the class of Forrester functions for 6 different methods. Solid lines display average over all the runs of given method, transparent areas depict $2$ standard deviations.}
\label{fig:1d_comp}
\end{figure*}

Adaptive Bayesian linear regression (ABLR) is Bayesian linear regression with nonlinear basis functions $\boldsymbol{\phi}_{\z} (\x): \mathbb{R}^D \rightarrow \mathbb{R}^d$, here parameterized by $\z$ ~\citep{bishop2006pattern}. \citet{ablr_perrone2018scalable} extend this approach to the multi-task setting by maintaining a separate Bayesian linear regression head for each task on top of a \emph{shared} neural network to learn the nonlinear basis functions:
\begin{align}
\label{eq:ablr}
\begin{split}
p(\yt|\wt,\z,\beta_t) &= \N (\mathbf{\Phi}_t(\z) \wt, \beta_t^{-1}\I_{N_t}), \\
p(\wt|\alpha_t) &= \N (\nul, \alpha_t^{-1}\I_{d}),
\end{split}
\end{align}
where $\yt \in \mathbb{R}^{N_t}$ is the vector of observed evaluations of $f_t$, $\wt \in \mathbb{R}^d$ is the weight vector of the Bayesian linear regression head $t$, and $\Pt(\z) = \boldsymbol{\phi}_\z (\Xt) \in \mathbb{R}^{N_t \times d}$ the corresponding design matrix with $\Xt \in \mathbb{R}^{N_t \times D}$ containing the configurations associated to $\yt$. The identity matrices $\I_{N_t}$ and $\I_d$ are respectively of size $N_t \times N_t$ and $d \times d$, and the scalars  $\alpha_t >0 $ and $\beta_t >0 $ are precisions parameters (or inverse variances). The weight vectors $\{\wt\}_t$ can be integrated out analytically. The resulting log-marginal likelihood can then be maximized with respect to the network parameters $\z$, as well as the task precisions $\{\alpha_t\}_t$ and $\{\beta_t\}_t$.

The shared underlying neural network used in multi-head ABLR makes it a suitable for multi-task BO~\citep{swersky2013multi,springenberg2016bayesian,ablr_perrone2018scalable}. Let $\mathcal{D}= \big\{ D_t : D_t =\{(\x_{tn}, y_{tn})\}_{n=1}^{N_t} \big\}_{t=1}^{T}$ be the data collected while optimizing the set of black-box functions $\{ f_t\}_{t=1}^T$. The basis functions $\boldsymbol{\phi}_\z(\cdot)$ encode a shared representation based on $\mathcal{D}$, while the task-specific head model parameters are learned from the corresponding $\mathcal{D}_t$. Hence, the optimization of a new task $f_{T+1}$ can be warm-started by adding a new head model on top of the neural network (that was trained  on $\mathcal{D}$) and use it as a surrogate model to optimize:
\begin{equation}
    \label{eq:bo_new}
    \x^{\star}_{T+1} = \argmin_{\x \in \mathcal{X}} f_{T+1}(\x),
\end{equation}
A practical advantage of multi-head ABLR is that the complexity of learning the neural network is linear in the total number of evaluations, unlike multi-ouput GPs~\citep{nguyen2014collaborative}, while the complexity of learning the head parameters is linear in the number of task evaluations. These attractive computational properties are retained by our proposed method, ABRAC, as discussed in more detail in Section~\ref{sec:abrac}.

\begin{algorithm}[tb]
   \caption{ABRAC}\label{alg:abrac}
\begin{algorithmic}
   \STATE {\bfseries Input:} number of initial points $n_0$, budget $N$, feature net $\phi_z(\cdot): \mathbb{R}^P \rightarrow \mathbb{R}^d$, filter $F^k$, previous evaluations $\{\{(x_{ti}, y_{ti})\}_{i=1}^{N_t}\}_{t=1}^T$.
   \STATE Fit $\phi_\z(\cdot)$ using $\{\{(x_{ti}, y_{ti})\}_{i=1}^{N_t}\}_{t=1}^T$ --Section~\ref{sec:offline_know_extr}.
   \STATE Observe $f_{\textit{new}}$ at $n_0$ randomly selected points $x_1, x_2, \dots, x_{n_0}\in \mathcal{X}$.
   \STATE $\mathcal{C} = \{(x_i, y_i)\}_{i=1}^{n_0}$, where $y_i = f(x_i)$.
    \STATE Set $n = n_0$.
   \WHILE{$n < N $}
   \STATE Fit probabilistic model $g$ based on  $\mathcal{C}$--Section~\ref{sec:online_tr}.
   \STATE $x_n = \argmax_{x \in \mathcal{X}} A_g(x)$, where $A$ is given acquisition function.
   \STATE Observe $y_n = f_{\textit{new}}(x_n)$. 
   \STATE Update $\mathcal{C} = \mathcal{C} \cup \{(x_n, y_n)\}$.
    \STATE $n = n + 1$ 
   \ENDWHILE
   \STATE {\bfseries Output:} $\hat{x} = \argmin_{i = 1,2,\dots, N} f_{\textit{new}}(x_i)$
\end{algorithmic}
\end{algorithm}

\begin{figure*}[t]
  \centering
  \subfigure[\texttt{ABRAC}]{\includegraphics[width=0.22\textwidth]{plots/1d_nest_drop.png}}
\hfill
  \subfigure[\texttt{ABLR L-BFGS}]{\includegraphics[width=0.22\textwidth]{plots/1d__ablr.png}}
\hfill
  \subfigure[\texttt{ABLR L-BFGS fixed}]{\includegraphics[width=0.22\textwidth]{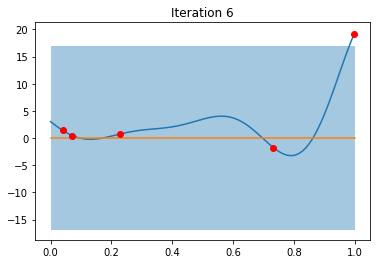}}
\hfill
  \subfigure[\texttt{ABLR SGD fixed}]{\includegraphics[width=0.22\textwidth]{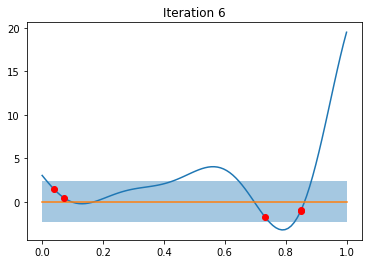}}
\hfill
  \caption{ Comparison of the surrogate objective along with its basis functions after 5 total evaluations obtained by 4 different methods. The target function is shown in blue, the evaluations are the red points (which includes the same $3$ initial random points for both methods), and the surrogate predictive mean is shown in orange with two standard deviations in transparent light blue.}
  \label{fig:1d_unc}
\end{figure*}

\section{Nested Dropout for Ordered Representations}\label{sec:nested}

Nested dropout is a structured type of dropout originally proposed for autoencoders to learn ordered representations in their bottleneck layer~\citep{nested_rippel2014learning}. It was also used in later works to learn spectral representations in convolutional neural networks~\citep{rippel2015spectral} and ordered word representations~\citep{liu2016learning}.

Assume there are $d$ neurons in the bottleneck layer. Nested dropout is applied during training by deactivating these neurons according to a discrete distribution $Q$ over the indices $1, \dots, d$. In each training step, one samples $b \sim Q$ and drops the neurons $b+1, \dots, d$. This results in the activation of a different, but nested subset of indices at each training step for which, if $b$ belongs to the subset, then $1, \dots, b-1$ also belongs to it. Hence, every mini-batch gradient update during training can only touch the neurons in the bottleneck layer that were not dropped, which, as a result, leads to an importance ranking of the representation dimensions. Nested dropout comes with strong theoretical properties for linear autoencoders: the solution obtained when applying nested dropout to the linear autoencoder recovers principal component analysis under an additional orthonormality constraint when $Q(b = j) > 0,  \forall j \in 1, \dots, d$~\citep[see][for details]{nested_rippel2014learning}.

\section{Multi-head ABLR with Adaptive Complexity (ABRAC)}\label{sec:abrac}

A deficiency of multi-head ABLR when applied to multi-task BO is that the number of nonlinear basis functions is not adapted to the target task, where the number of observations is typically smaller than in the previous tasks, making it prone to overfitting. This issue is due to the direct use of conventional multi-task models in the context of sequential decision making problems such as BO. To solve this issue we propose to learn an ordered set of learned basis functions, which we selectively activate during the transfer learning procedure. Pseudocode is given in Algorithm~\ref{alg:abrac}.

To endow multi-head ABLR with ordered representations for transfer learning, we apply nested dropout during training to the last layer of the shared neural network $\boldsymbol{\phi}_\z (\x)$, which operates as the feature extractor. At each training step, we activate a nested subset of the output neurons according to a uniform distribution over the indices: $Q(b = j) = \nicefrac{1}{d}$, $\forall j \in 1, \dots, d$. Departing from \citet{nested_rippel2014learning}, who proposed to sample the truncation index according to a geometric distribution, we observed empirically that the uniform distribution performed well without any further modifications, and with the additional benefit that we sample neurons with a high index more often and do not have to set additional hyperparameters.

Next, we introduce multi-head ABLR with adaptive complexity (ABRAC) for multi-task BO and detail the inference procedure. We propose a two-step approach to do transfer learning in the context of Bayesian sequential decision making consisting in (1) an off-line learning of ordered shared representations and (2) an online adaptation of the transfer learning procedure to improve the sample efficiency of the target BO task.

\textbf{Probabilistic model.} We make a minor modification of (\ref{eq:ablr}), which consists in allowing for a different precision parameter per basis function:
\begin{align}
\label{eq:model}
p(\wt | \boldsymbol{\alpha}_t) &= \N (\nul, \Diag(\boldsymbol{\alpha}_t)^{-1}),
\end{align}
where $\alpha_{ti}>0$ is the precision associated to basis function $i$. Training consists in optimizing the log-marginal likelihood wrt the shared parameters $\z$ and the head model parameters $\{\boldsymbol{\alpha}_t,\beta_t\}_t$:
\begin{align}
\label{eq:logmarginal}
&\sum_t\ln p(\yt | \z, \boldsymbol{\alpha}_t,\beta_t) =\\
&\sum_t\ln \N(\nul , \beta_t^{-1}\I_{N_t} + \Pt(\z)\Diag(\boldsymbol{\alpha}_t)^{-1}\Pt(\z)^\top) . \nonumber\
\end{align}
Allowing for different $\alpha_{ti}$ promotes sparsity by enabling us to automatically determine the relevance of every single component of $\wt$~\citep{mackay}. The application of automatic relevance determination (ARD) to Bayesian linear regression was studied in \citep{rvm2} and \citep{wipf}.

\begin{figure*}[t]
\centering
\includegraphics[width=0.31\textwidth]{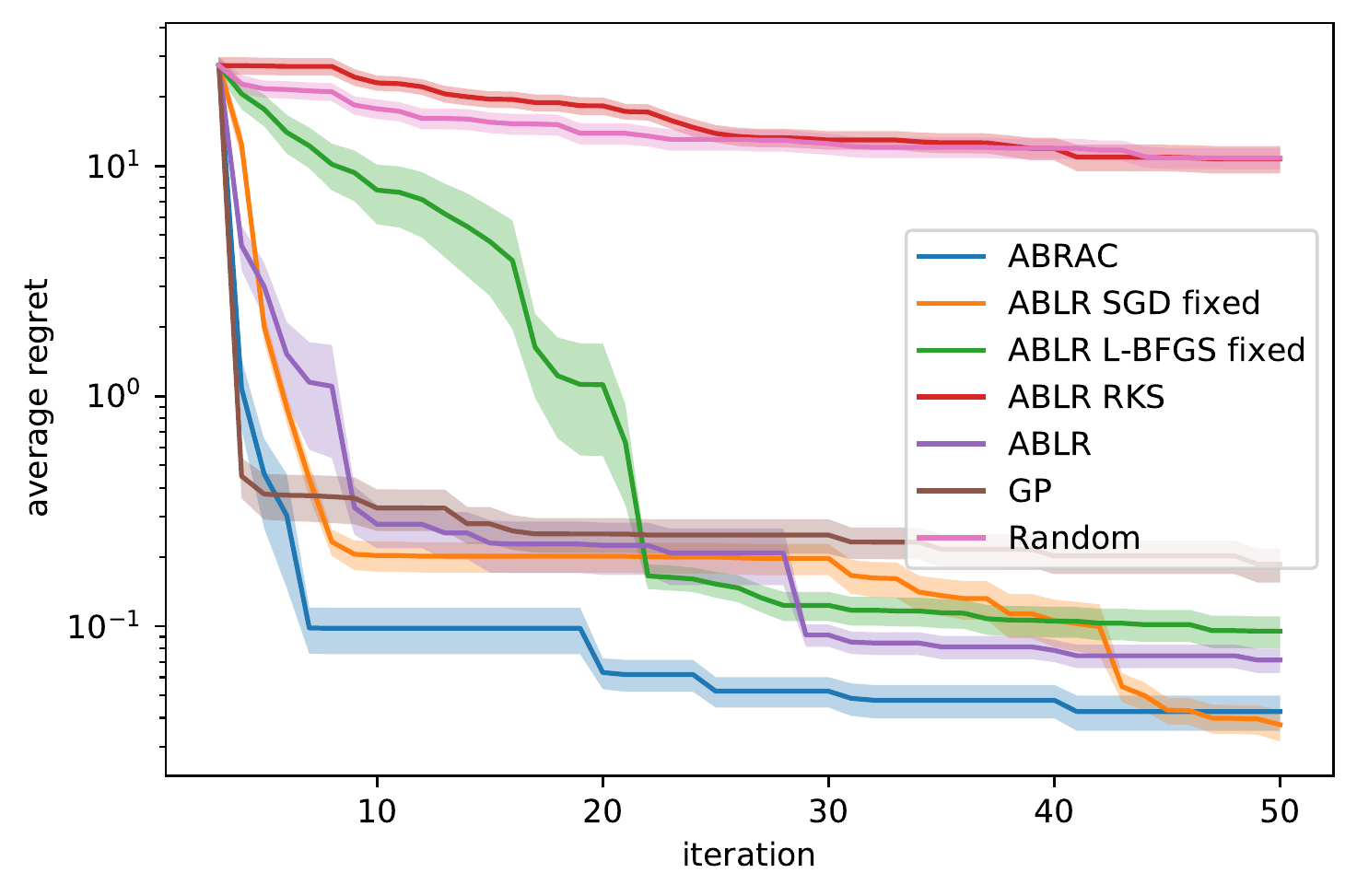}
\includegraphics[width=0.31\textwidth]{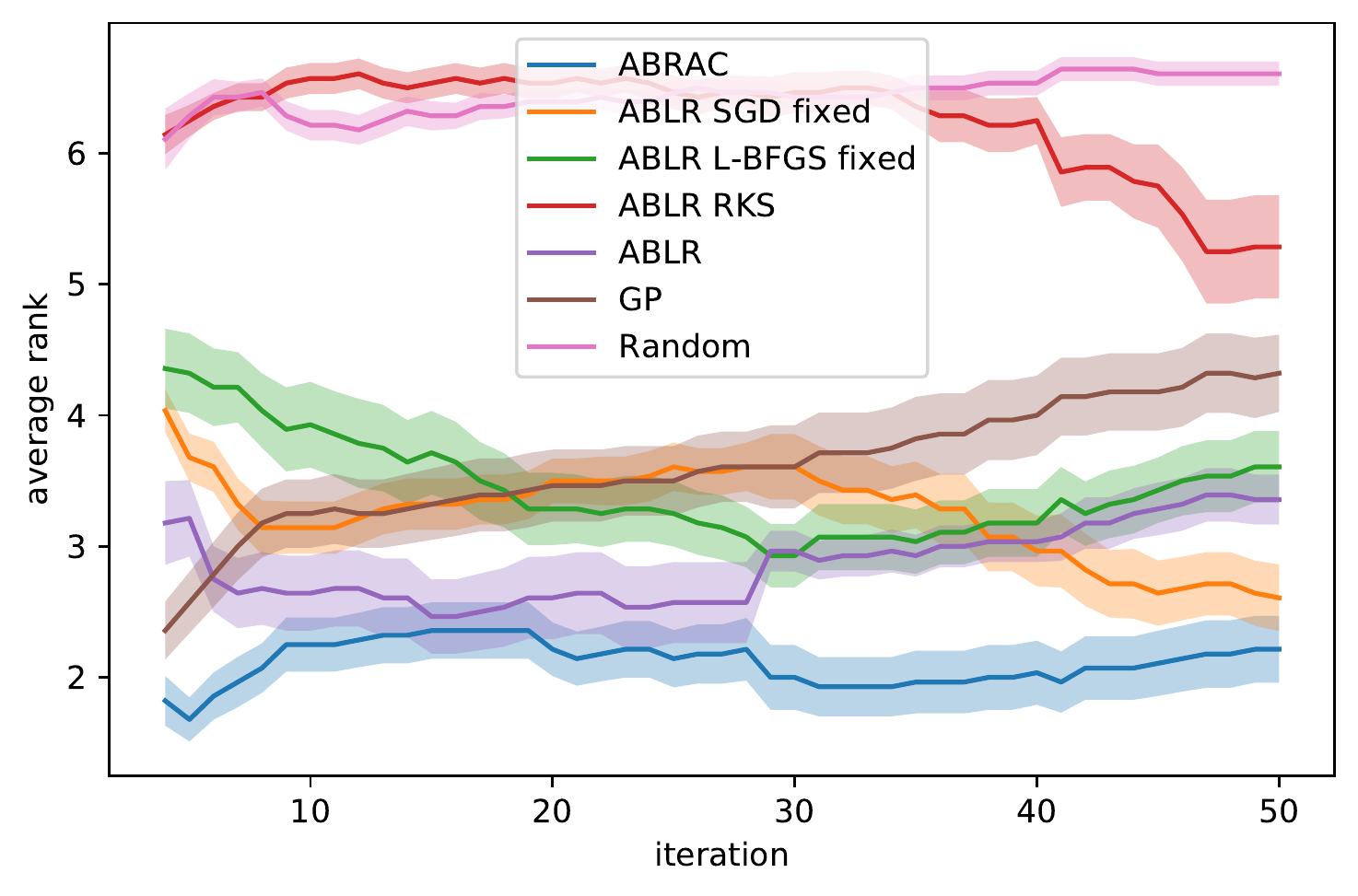}
\includegraphics[width=0.31\textwidth]{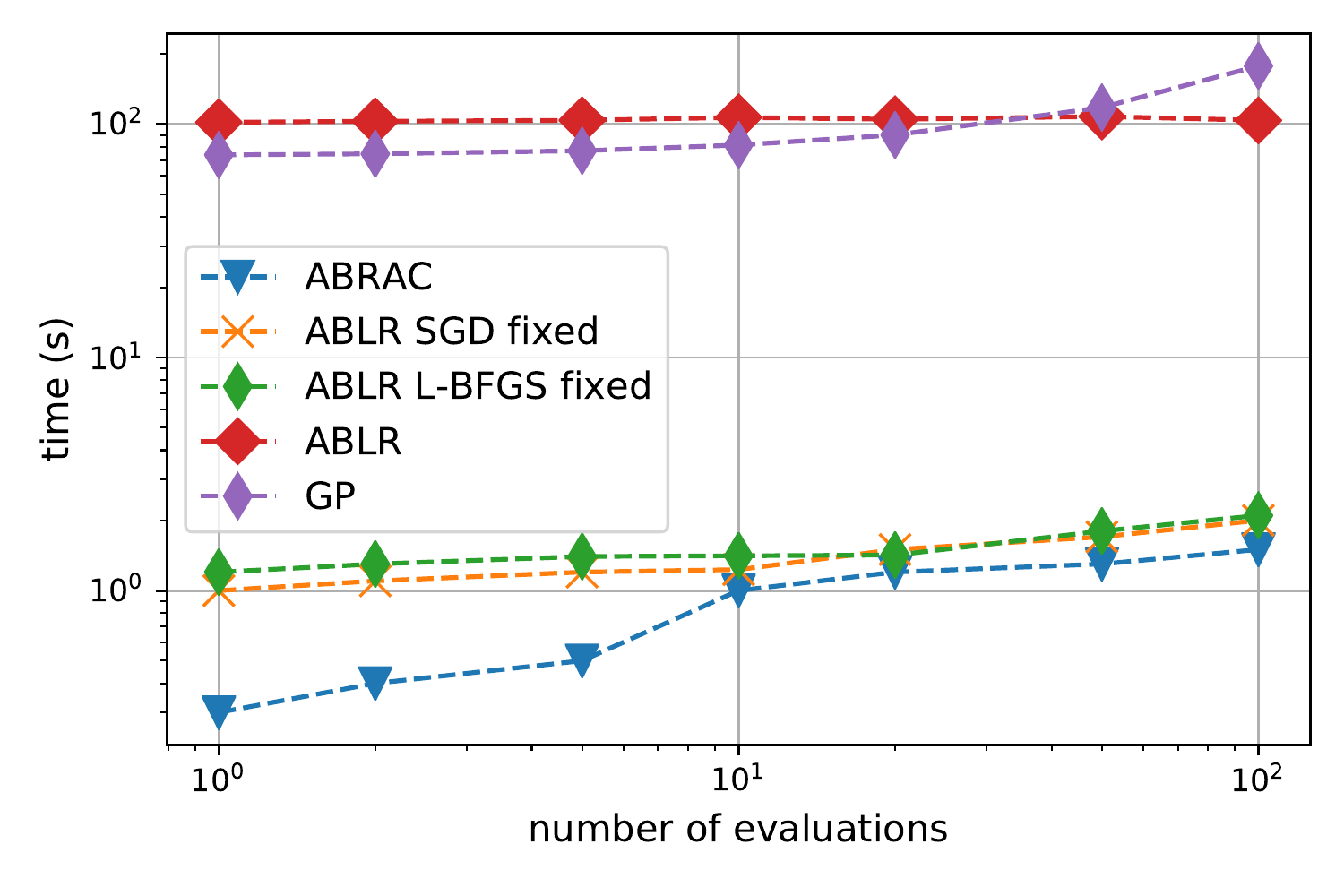}
\caption{Comparison of the average regret and rank on the class of parameterized quadratics for 7 different methods. Solid lines display average over all the runs of given method, transparent areas depict $2$ standard deviations. The most right plot displays comparison of the speed of ABRAC, different ABLR based methods and GP enriched by subsampling.}
\label{fig:quad_comp}
\end{figure*}

\subsection{Offline Ordered Representation Learning}
\label{sec:offline_know_extr}

To learn ordered representations suitable for transfer learning, we first train a neural network $\phi_\z(\cdot)$ off-line. We minimize the mean square error between neural network predictions and the observations collected during all previous tasks using SGD with momentum~\citep{sutskever2013importance}. This is different from \citet{ablr_perrone2018scalable}, where L-BFGS is used to learn $\z$ jointly with $\{\boldsymbol{\alpha}_t,\beta_t\}_t$, and akin DNGO~\citep{dngo_snoek2015scalable}. However, we apply nested dropout to the last layer during training as discussed next.

Before applying a mini-batch update, we first sample a random truncation $b$ for each data in the mini-batch. Let $\boldsymbol{\phi}(\x)_{b\downarrow}$ be a random truncation of the vector $\boldsymbol{\phi}(\x)$ where the last $d-b$ elements are masked with 0. For example, $\boldsymbol{\phi}(\x)_{2\downarrow}^\top = [\phi_1(\x),\phi_2(\x),\phi_3(\x),\phi_4(\x)]_{2\downarrow}^\top = [\phi_1(\x),\phi_2(\x), 0, 0]^\top$. If $b = d$, then $\boldsymbol{\phi}(\x)_{b\downarrow}=\boldsymbol{\phi}(\x)$ and $(\cdot)_{b\downarrow}$ is the identity operator. Hence, we can view the off-line training procedure as maximizing a likelihood of which the terms correspond to a product of normal distributions:
\begin{align}
\label{eq:modelinterpret}
p(y_{tn}|\wt,\z,\beta_t) &= \prod_{i=1}^d \N (\boldsymbol{\phi}_{\z}(\x_{tn})_{i\downarrow}^\top\wt, \beta_t^{-1})^{\delta_{ib_{tn}}} ,
\end{align}
where $\delta_{ij}$ is the Kronecker delta and the truncation $b_{tn}\sim \mathrm{Uniform}(1,\ldots,d)$ is resampled at each iteration of the optimizer.

\subsection{Online Adaptive Transfer Learning}
\label{sec:online_tr}

When tackling a new BO task, we fix the feature extractor $\boldsymbol{\phi}_\z(\x)$ and learn the additional head model parameters by maximizing the corresponding term in (\ref{eq:logmarginal}) after each addition of a new observation. Doing so dynamically adjusts the level of sparsity of Bayesian linear regression (BLR) via ARD. Maximizing the log-marginal likelihood of BLR has overall complexity $\mathcal{O}(d^2\max\{N_\n, d\})$, which is linear in number of evaluations of the new BO task. See \citet{bishop2006pattern}.

\section{Experiments}\label{sec:experiments}

We now present several experiments that show the benefits of our new proposed method, dubbed ABRAC.
Firstly, we use the 1-D synthetic Forrester function, which allows us to visually quantify the model fit and uncertainty estimations.
Afterwards, we present results on the parametrized quadratic functions by~\citet{ablr_perrone2018scalable}.
Lastly, we benchmark ABRAC on real-world HPO problems containing feed-forward neural networks and support vector machines.

We implemented ABRAC and ABLR in the python framework Emukit~\citep{emukit2018} where the neural network code is based on MXNet~\citep{mxnet_chen2015mxnet}. For L-BFGS, we used the SciPy~\citep{scipy} implementation.
For Gaussian Process (GP) based BO,  we used a Matern52 to model the correlation between hyperparameters with automatic relevance determination, and optimize all hyperparameters by empirical Bayes~\citep{rasmussen2003gaussian}.  
All BO methods used expected improvement as acquisition function~\citep{BO_mockus1978application}.  

\begin{figure*}[!t]
\centering
\includegraphics[width=0.39\textwidth]{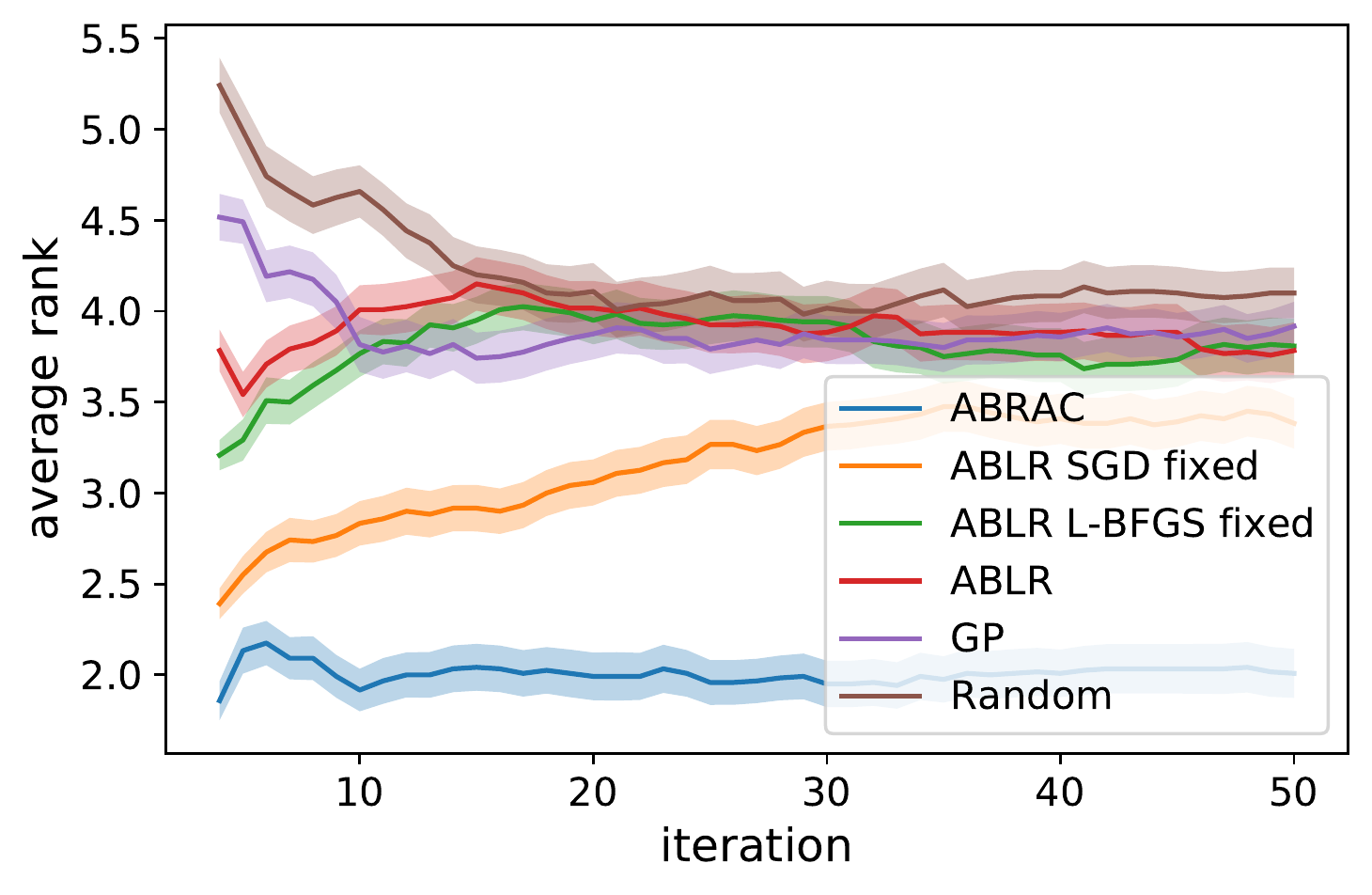}
\caption{Comparison of the average rank on the NAS Benchmarks for 5 different methods. Solid lines display average over all the runs of given method, transparent areas depict $2$ standard deviations. }
\label{fig:nas_comp}
\end{figure*}

\begin{figure*}[!t]
  \centering
 \includegraphics[width=0.24\textwidth]{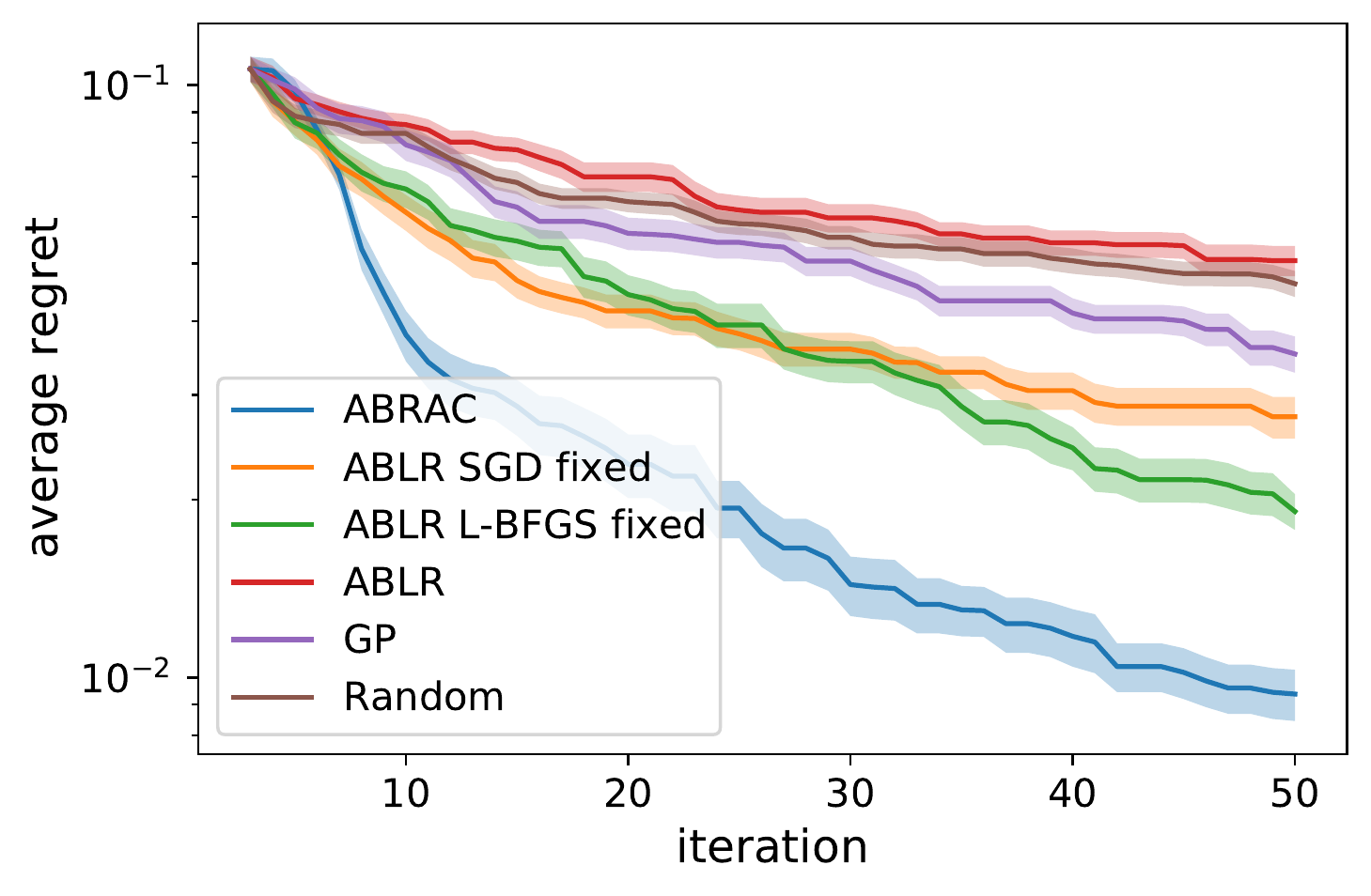}
  \includegraphics[width=0.24\textwidth]{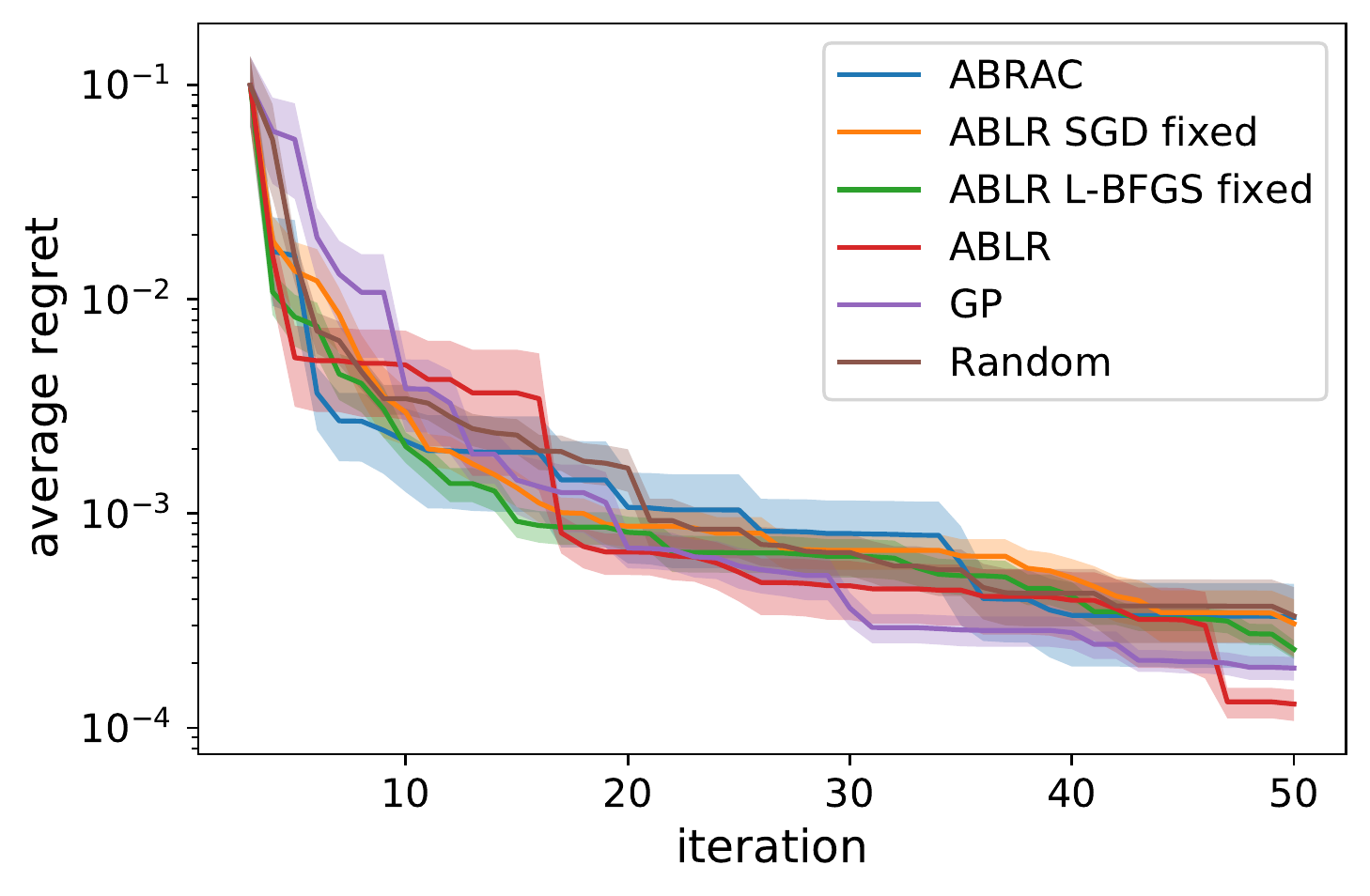}
   \includegraphics[width=0.24\textwidth]{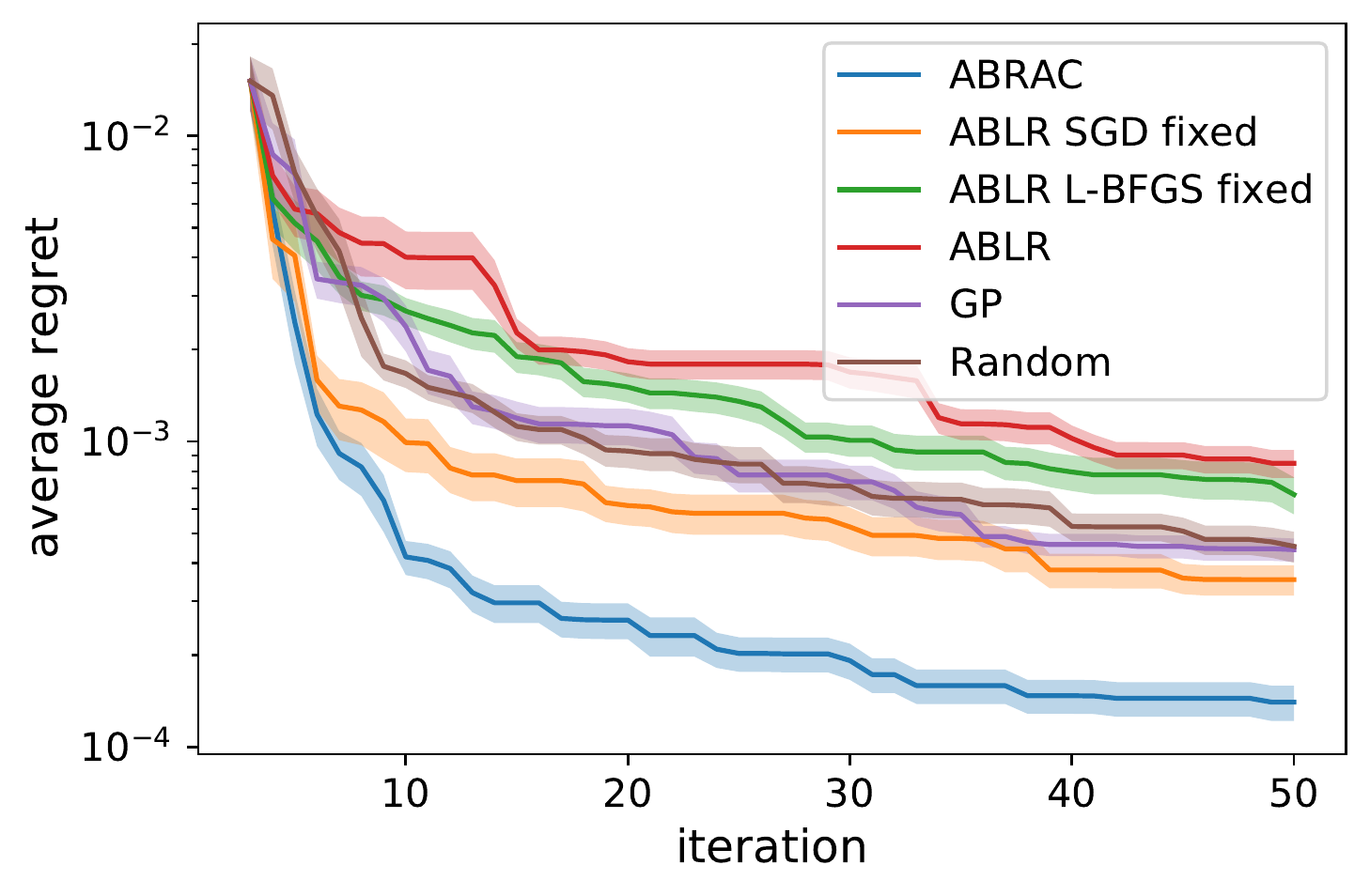}
  \includegraphics[width=0.24\textwidth]{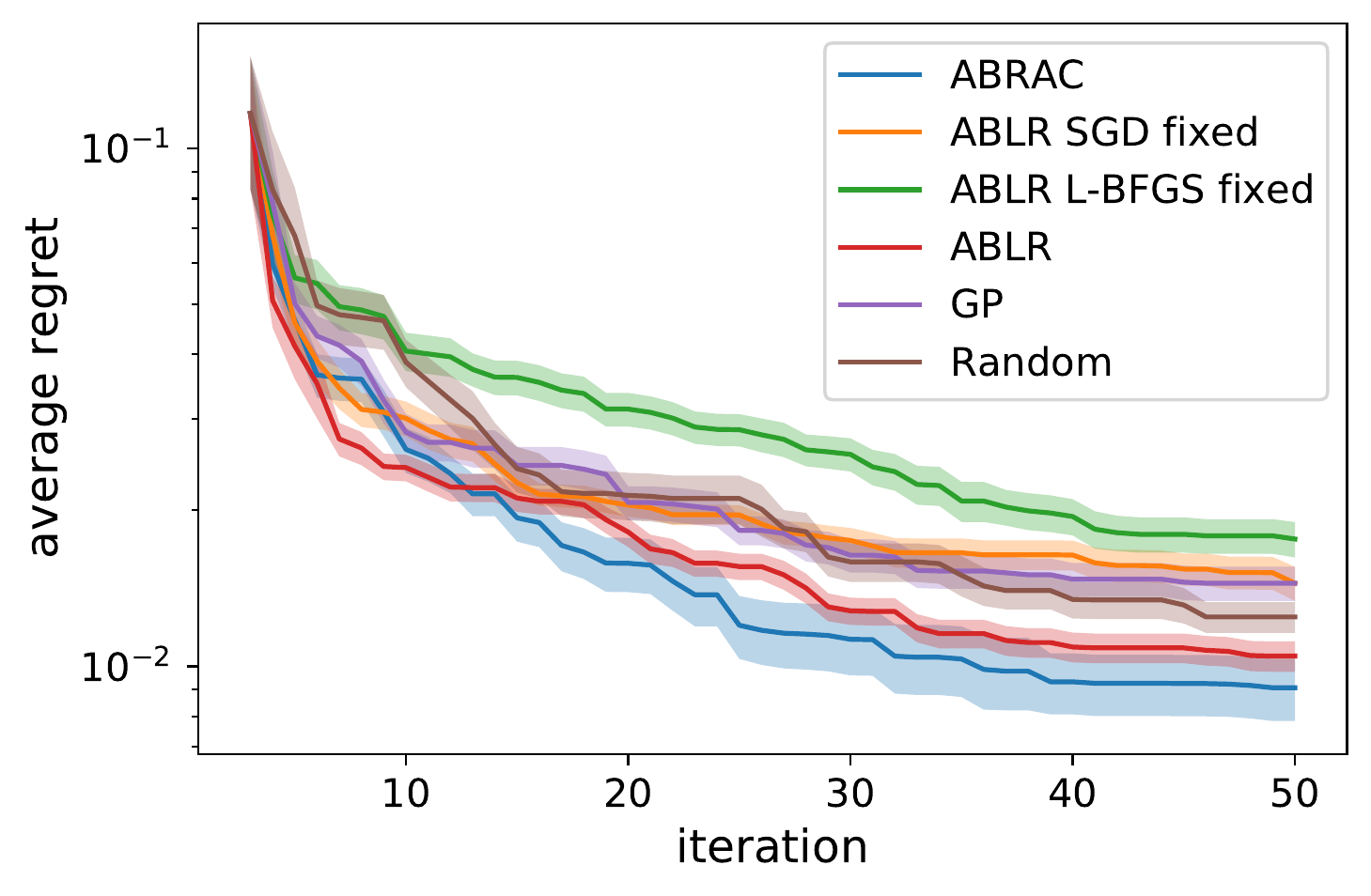}
\hfill
 
   \includegraphics[width=0.24\textwidth]{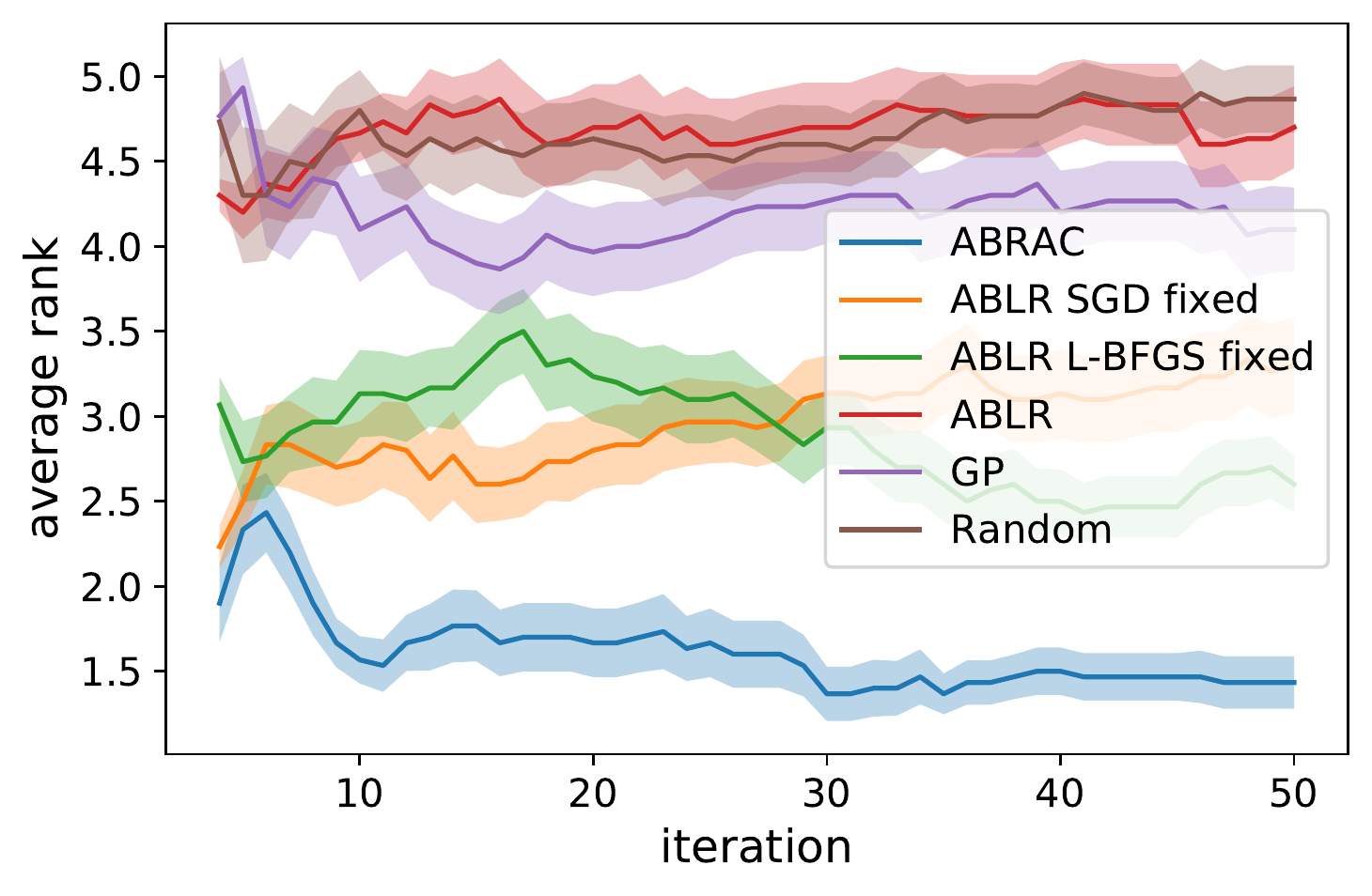}
  \includegraphics[width=0.24\textwidth]{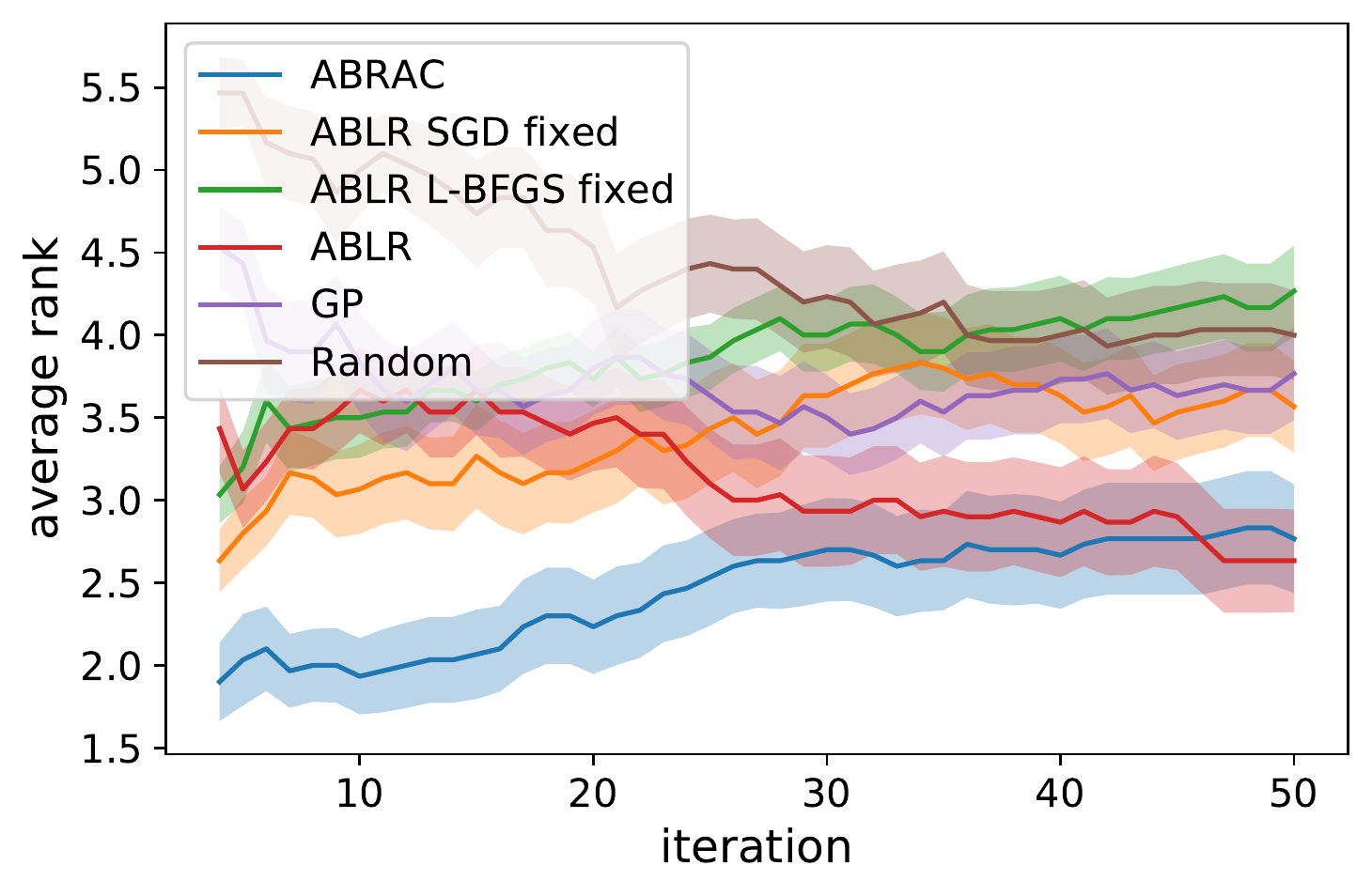}
   \includegraphics[width=0.24\textwidth]{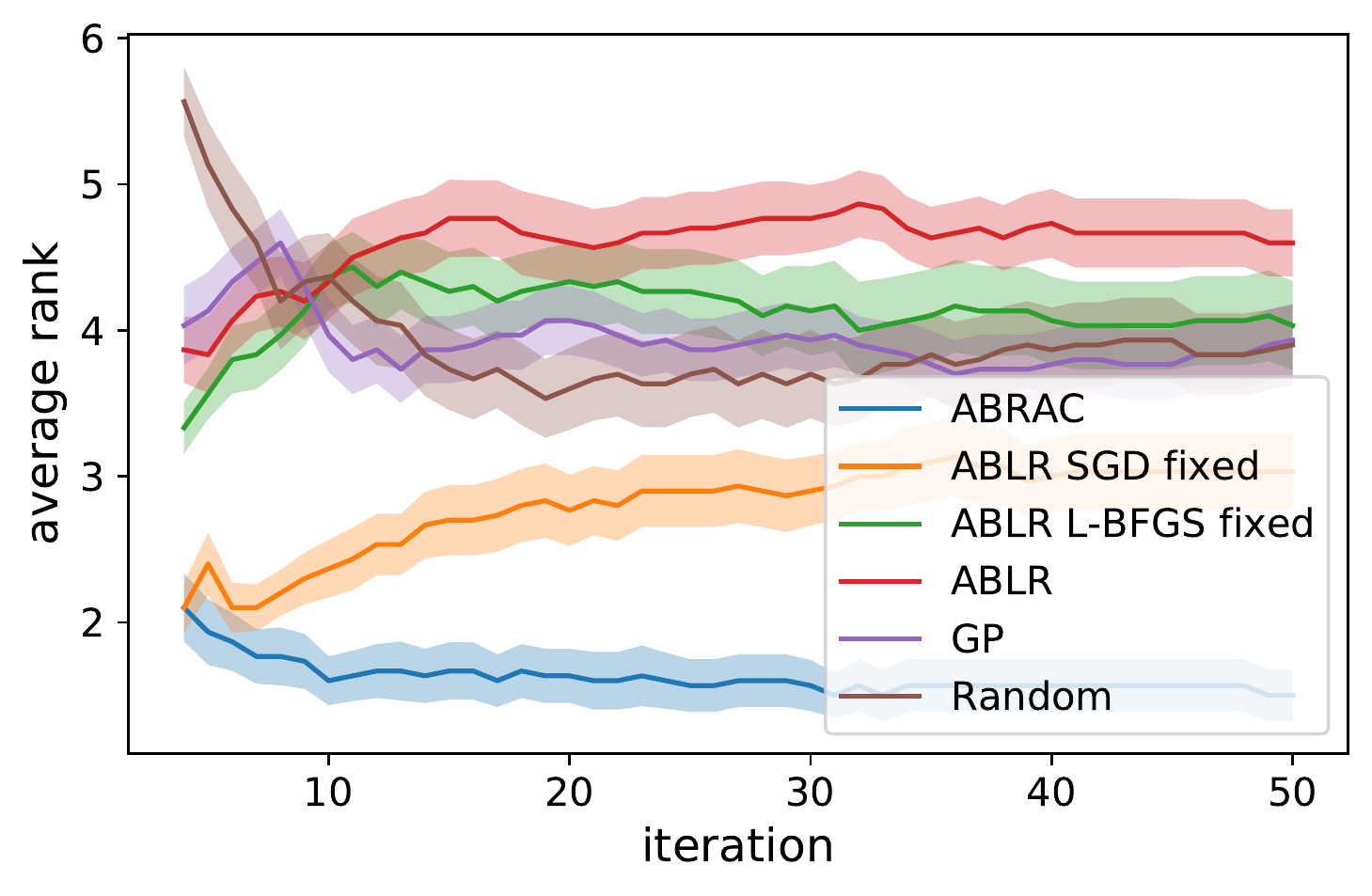}
  \includegraphics[width=0.24\textwidth]{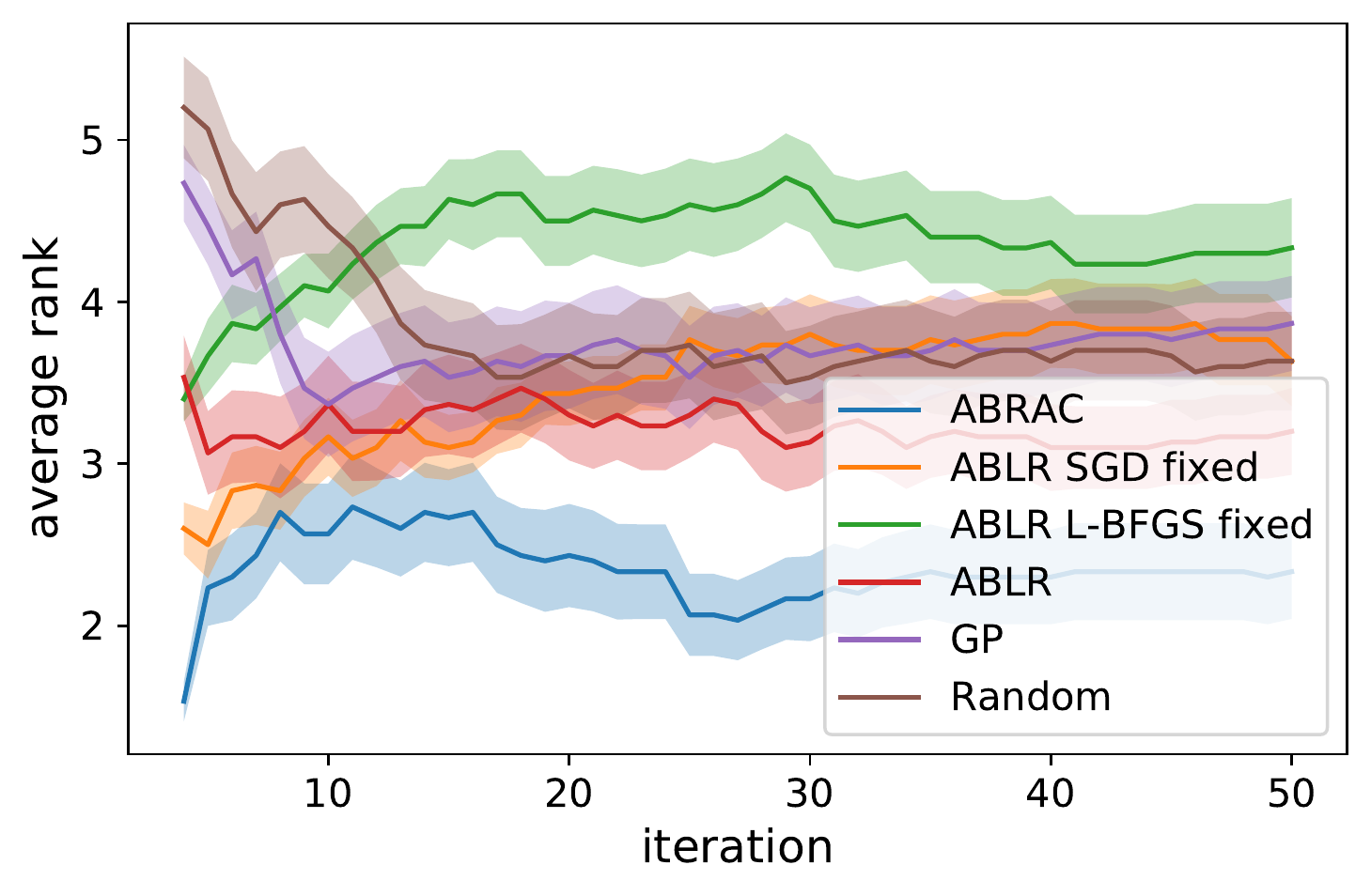}
\hfill
\caption{Comparison of the average regret and rank on the NAS Benchmarks for 5 different methods. Each column corresponds to one dataset (from left:  \texttt{Protein Structure, Naval Propulsion, Slice Localization}, and \texttt{Parkinsons Telemonitoring}). Solid lines display average over all the runs of given method, transparent areas depict $2$ standard deviations. 
}
  \label{fig:nas_ind}
\end{figure*}

We include BOHAMIANN~\citep{springenberg2016bayesian} as baseline, but, because of its significant computational overhead (approximately $100\times$ slower than ABLR~\citep{ablr_perrone2018scalable}) we show its performance only on the first experiment. For all other experiments, we ran BOHAMIANN for $10$ iterations and it did not outperform ABRAC in any of those.

To highlight the benefits of using an offline-trained feature extractor with hierarchical basis functions, we include the following two modifications of ABRAC without nested dropout: \textit{ABLR SGD fixed} trains the features extractor by minimizing the squared error with SGD, which can be seen as an efficient multi-task version of DNGO~\citep{dngo_snoek2015scalable} and \textit{ABLR L-BFGS fixed} which minimizes the negative log-likelihood with L-BFGS. Unlike  \textit{DNGO}, both versions do not require any contextual information. The main difference to standard \textit{ABLR} is that the neural network $\boldsymbol{\phi}_{\z}$ is fixed for the new task.
Furthermore, to analyse the benefit of using a neural network based feature extractor, we also include another baseline that uses random kitchen sinks as basis functions (ABLR RKS).

For each experiment, we average the performance across tasks in a leave-one-task-out fashion.
Each optimization run was initialized with $3$ random data points and we reported average performance of each method across $5$ repetitions with a different seed.
All experiments were conducted on AWS \texttt{ml.c4.2xlarge} instances.

\begin{figure*}[t]
\centering
\includegraphics[width=0.37\textwidth]{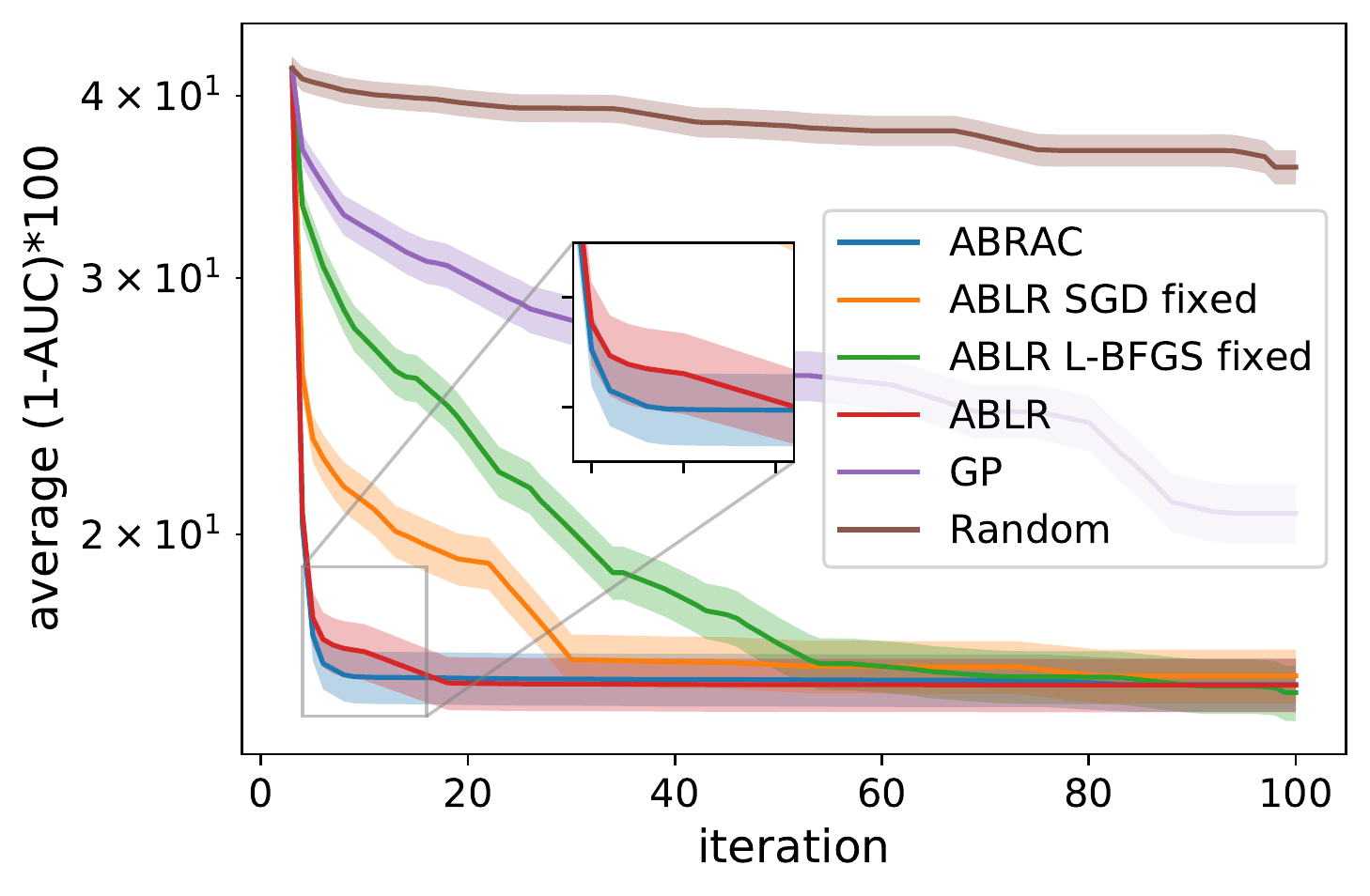}
\includegraphics[width=0.37\textwidth]{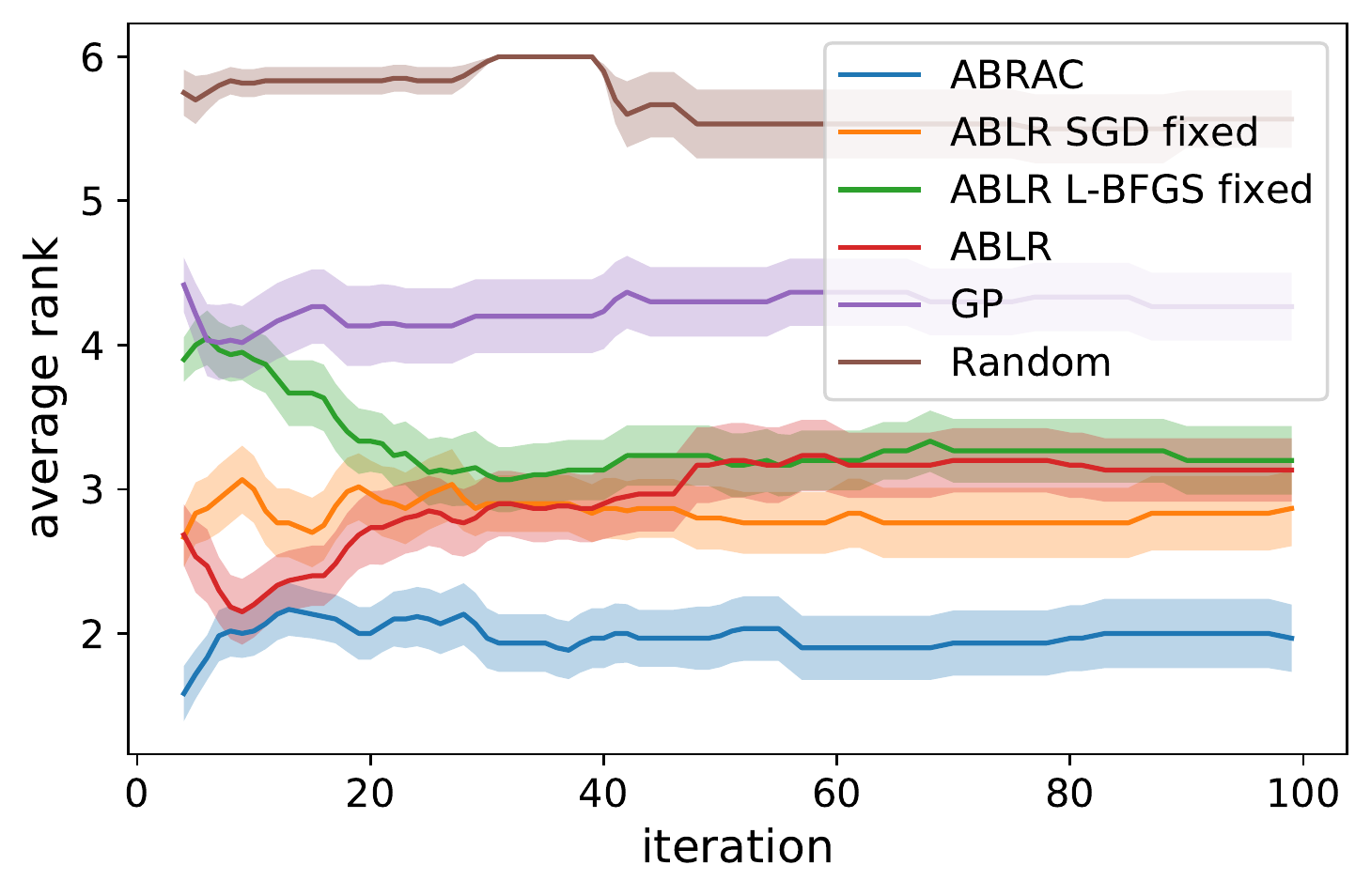}
\caption{Comparison of the average AUC and rank on the OpenML-SVM Benchmarks for 6 different methods. Solid lines display average over all the runs of given method, transparent areas depict $2$ standard deviations.}
\label{fig:openmlsvm_comp}
\end{figure*}

\textbf{Forrester Function.}
For the first experiment, we consider the parameterized class of Forrester functions $f_t(x|a_t,b_t,c_t) = (a_t x-2)^2\sin(b_t x-4)+c_t$ with the configuration space $\mathcal{X} = [0,1]$. We create new tasks by sampling different coefficients $a_t \sim \N(6,1)$, $b_t \sim \N(12,4)$, $c_t \sim \mathrm{Uniform}(0,10)$.
We generated 10 tasks with $20$ uniformly distributed data points each.
The architecture of the feature extractor $\boldsymbol{\phi}_\z(\x)$ consists of a $2$-layers fully connected neural network with $50$ hidden units and $\tanh$ activation and $d=20$ output units with linear activation functions.

Firstly, we investigate the effects of the two major components of ABRAC: the offline training with nested dropout and the online transfer learning with adaptive basis function selection. 
To address the former we compare against a modified version of ABLR trained with SGD, dubbed \textit{ABLR SGD adapt}, where we incorporate our adaptive procedure during the optimization. 
For  the latter we include a version of ABRAC which considers all basis functions at every step (ABRAC fixed).  
Looking at Figure~\ref{fig:ablation}, one can see that not only in isolation, each of the two components improves over vanilla ABLR (with SGD training) but in combination, i.e  ABRAC, we achieve the best performance. 

In Figure~\ref{fig:1d_comp} we compare against a range of baselines: Multi Task Gaussian process based Bayesian optimization (MT-GP-BO),  which is a standard GP defined over the search space and the task-specific contextual information $(a_t, b_t, c_t)$ , random search (RS), ABLR, BOHAMIANN as well as ABLR SGD fixed and ABLR L-BFGS fixed ascribed above.
Similarly to MT-GP-BO,  BOHAMIANN is  provided with contextual information $(a_t, b_t, c_t)$ for each task to enable multi-task learning.  No other algorithm is given this information.
One can see that ABRAC quickly adapts to the new task and achieves already after 5 function evaluations an acceptable regret. In contrast,  GP-BO and ABLR require more samples to approach the global optimum.
Interestingly, the baselines ABLR SGD fixed and ABLR L-BFGS fixed perform substantially worse, highlighting the importance to nested dropout to learn adaptive basis functions.
To obtain a better intuition of our method, we show in Figure~\ref{fig:1d_unc} the posterior mean and standard deviation of ABRAC and the ABLR variants after two optimization steps. All methods were initialized with the same three data points.
Vanilla ABLR is not able to account for the reduced complexity of this low data regime and heavily overfits.
On the other hand, ABLR with fixed basis functions explains all data points with observation noise. 
ABRAC is able to capture the right complexity and provides already in this early stage of optimization a meaningful model. Similar plots can be found in Figure~\ref{fig:1d_unc_app} in the Appendix, where we show the same plots but with 5 fixed data points among all the tasks. In addition, we also show the extracted basis functions with their relative weights in Figure~\ref{fig:1d_unc_intro}. 

\textbf{Parameterized Quadratic functions.}

In the second experiment, we compare different methods on the parameterized quadratic functions of the form $f_t(x|a_t,b_t,c_t) = a_tx^\top x + b_t \mathbf{1}^\top x + c_t$ from \citet{ablr_perrone2018scalable}, where $\mathbf{1}$ is an all-ones vector, and the coefficients are sampled uniformly $a,b,c \sim \mathrm{Uniform}(0.1, 1)$.
The configuration space is defined by $\mathcal{X} = [-10, 10]^5$.
To train the feature extractor we generated $30$ tasks with $100$ random data points each.

Due to scalability issues we warm started MT-GP-BO with only $10$ randomly selected datasets from each task, i.e in total $290$ data points. 
In order to distinguish between tasks, we provided MT-GP-BO with contextual information $(a_t, b_t, c_t)$ for each task.
Note that, none of the other methods is provided with contextual information.
Figure~\ref{fig:quad_comp} shows that ABRAC is able to outperform all other methods. The only method that can reach ABRAC's performance after $45$ iterations is ABLR SGD fixed. 
In Figure~\ref{fig:quad_comp} right, we report the optimization overhead of all neural network based methods.
One can see that in all scenarios, ABRAC is not just the most efficient method in terms of sampling efficiency, but also in terms of computational overhead where it is more than  $100 \times$ faster than ABLR or GPs.

\textbf{Tabular Benchmarks.}
For the next experiment, we used the tabular benchmark by \citet{klein-arxiv19a} which simulates the discrete HPO of a 2-layer fully-connected neural network on 4 different regression datasets.
\citet{klein-arxiv19a} performed an exhaustive search of all $62208$ possible configurations in the search space per dataset, such that for a single function evaluation one can simply look up the performance in the table instead of training the neural network from scratch.
For further details we refer to the original paper.
For each task, we generate  $1024$ random configurations which are used for the offline feature learning. Since we have only access to $4$ datasets, we increase the number of repetition with a different seed to $10$.  Since we have no access to contextual information we don't include MT-GP-BO as a baseline but only GP-BO.

Figure~\ref{fig:nas_comp} shows that in terms of rank, ABRAC outperfoms all other method with significant margin, which do not improve much upon random search in the given budget. We include 4 additional graphs in Figure~\ref{fig:nas_ind}, where each of the figures displays average regret and rank for one of four datasets. One can see that for all the datataset, ABRAC is the best in terms of average rank and the same holds for the average regret except for \texttt{Naval Propulsion}. However, for this dataset random search already achieves a good performance and none of the methods significantly improves over it.

\textbf{SVM on OpenML datasets.}

Lastly, we consider the HPO of support vector machines (SVM) with 2 continuous hyperparameters: the regularization parameter \textit{C}(min: 0.000986, max: 998.492437) and the kernel parameter \textit{gamma}(min: 0.000988, max: 913.373845) on OpenML~\citep{vanschoren2014openml} classification datasets.
We chose the most evaluated flow (\texttt{flow id} 5891) with its 30 most evaluated datasets/tasks with almost one million evaluations in total.

We provide MT-GP-BO with contextual information: the number of features, the number of data points and the percentage of the majority class, and warm start it using 10 random points from each of the 29 related tasks.
We compare these methods based on the average rank and average Area Under Curve (AUC). To model AUC on unseen configurations, we use the surrogate modeling approach from \citet{eggensperger2015efficient} based on a random forest model. 
Figure~\ref{fig:openmlsvm_comp} confirms the superior behaviour of ABRAC to all other methods while the only method that can reach comparable performance in terms of AUC is ABLR, which considers all data in each step. 

\section{Conclusion}
\label{sec:disc}

We introduced ABRAC, a probabilistic model for Bayesian optimization with adaptive complexity.
It first learns a set of basis function based on neural networks on offline generated data with increasing complexity.
During Bayesian optimization, we use Bayesian linear regression to obtain uncertainty estimates and thereby regularizes our basis functions.
Compared to Gaussian processes our model scales linearly rather than cubically with the number of datapoints.
In a battery of experiments, we show that ABRAC obtains much higher speed ups compared to other baseline methods, such as ABLR or Gaussian processes based Bayesian optimization.

\bibliography{references,lib}

\onecolumn
\appendix

\begin{figure*}[t]
  \centering
  \subfigure[\texttt{ABRAC}]{
  \includegraphics[width=0.24\textwidth]{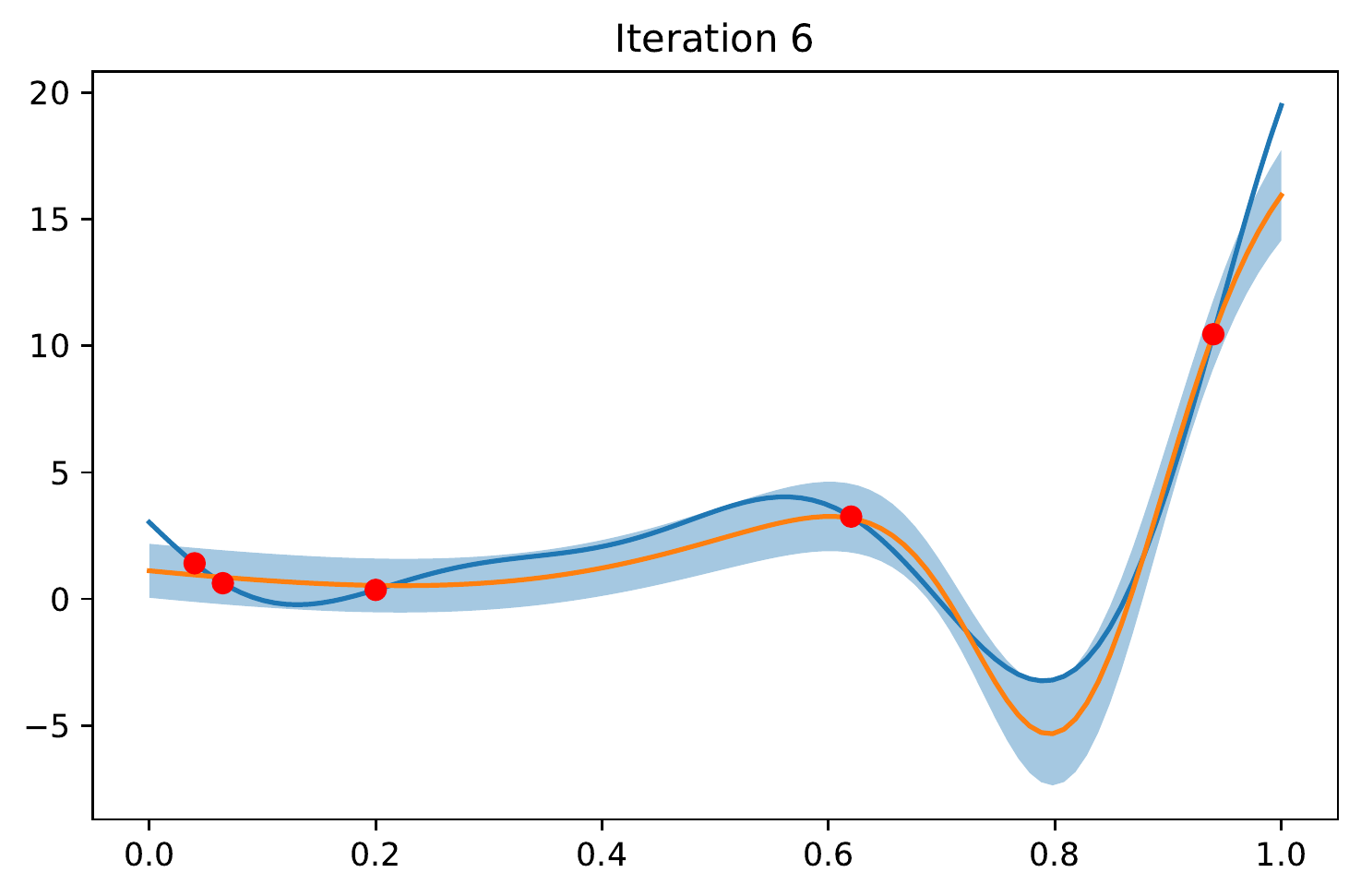}
  \includegraphics[width=0.24\textwidth]{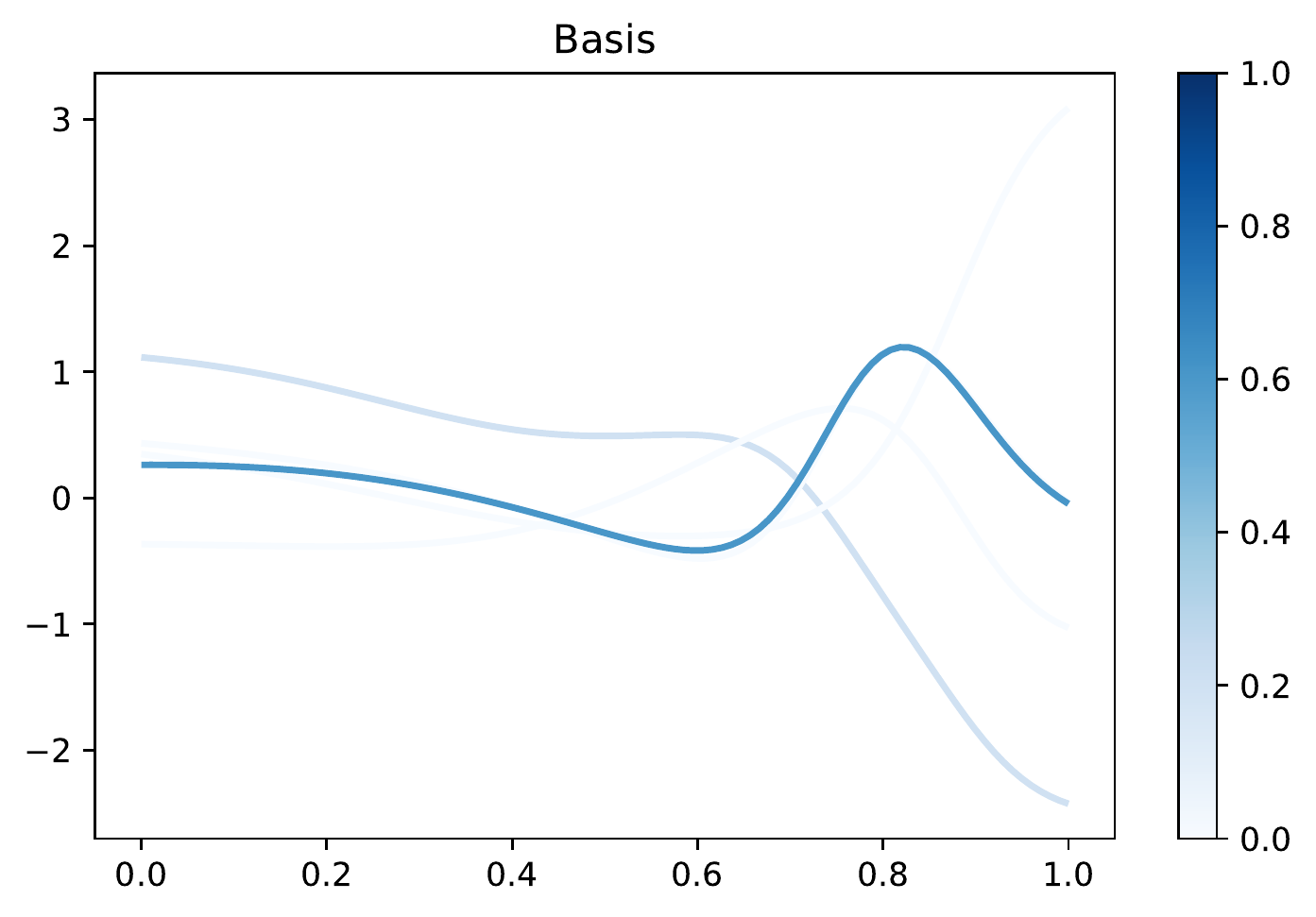}}
\hfill
  \subfigure[\texttt{ABLR}]{
  \includegraphics[width=0.24\textwidth]{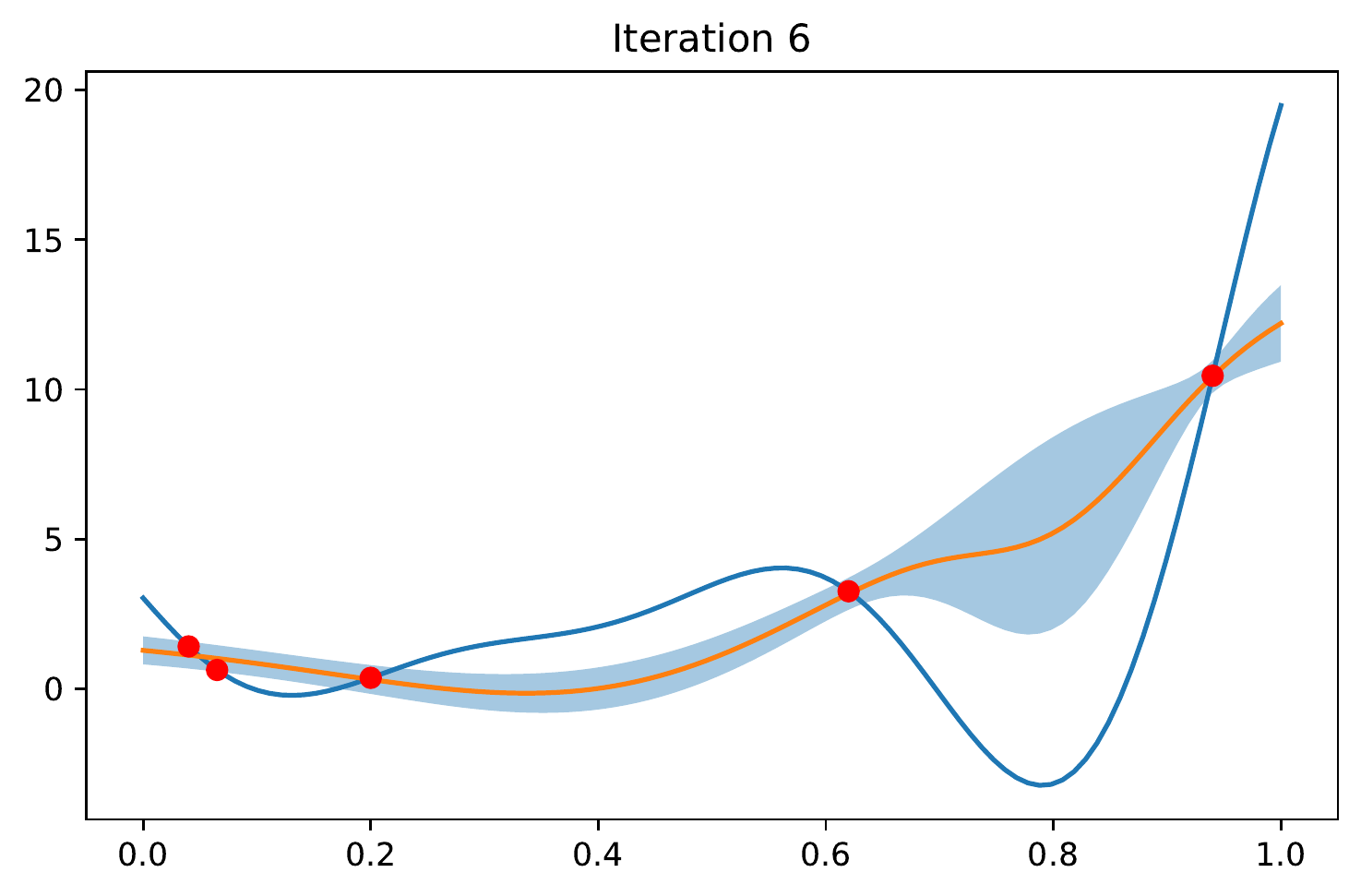}
  \includegraphics[width=0.24\textwidth]{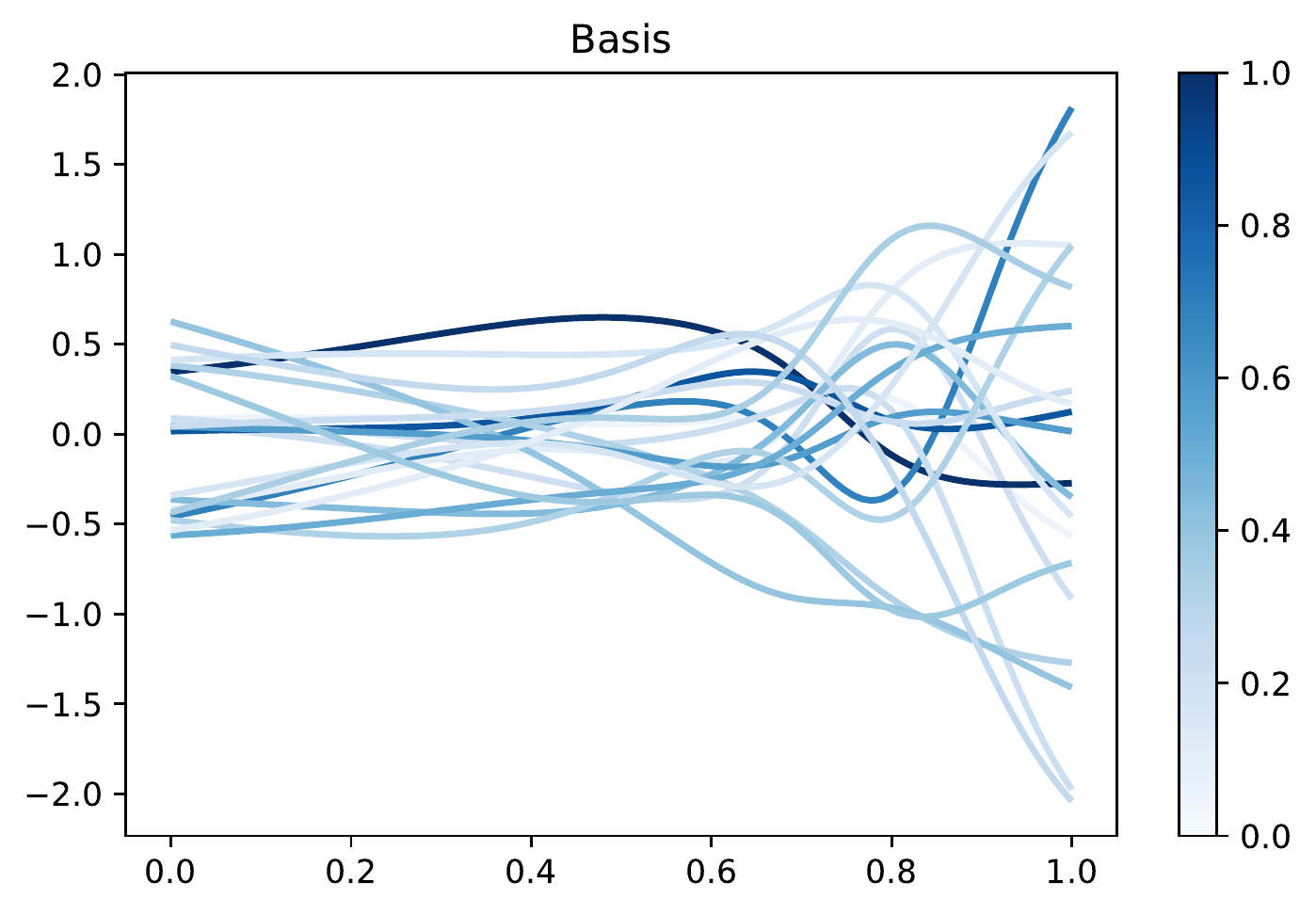}}
\hfill
  \subfigure[\texttt{ABLR SGD fixed}]{
  \includegraphics[width=0.24\textwidth]{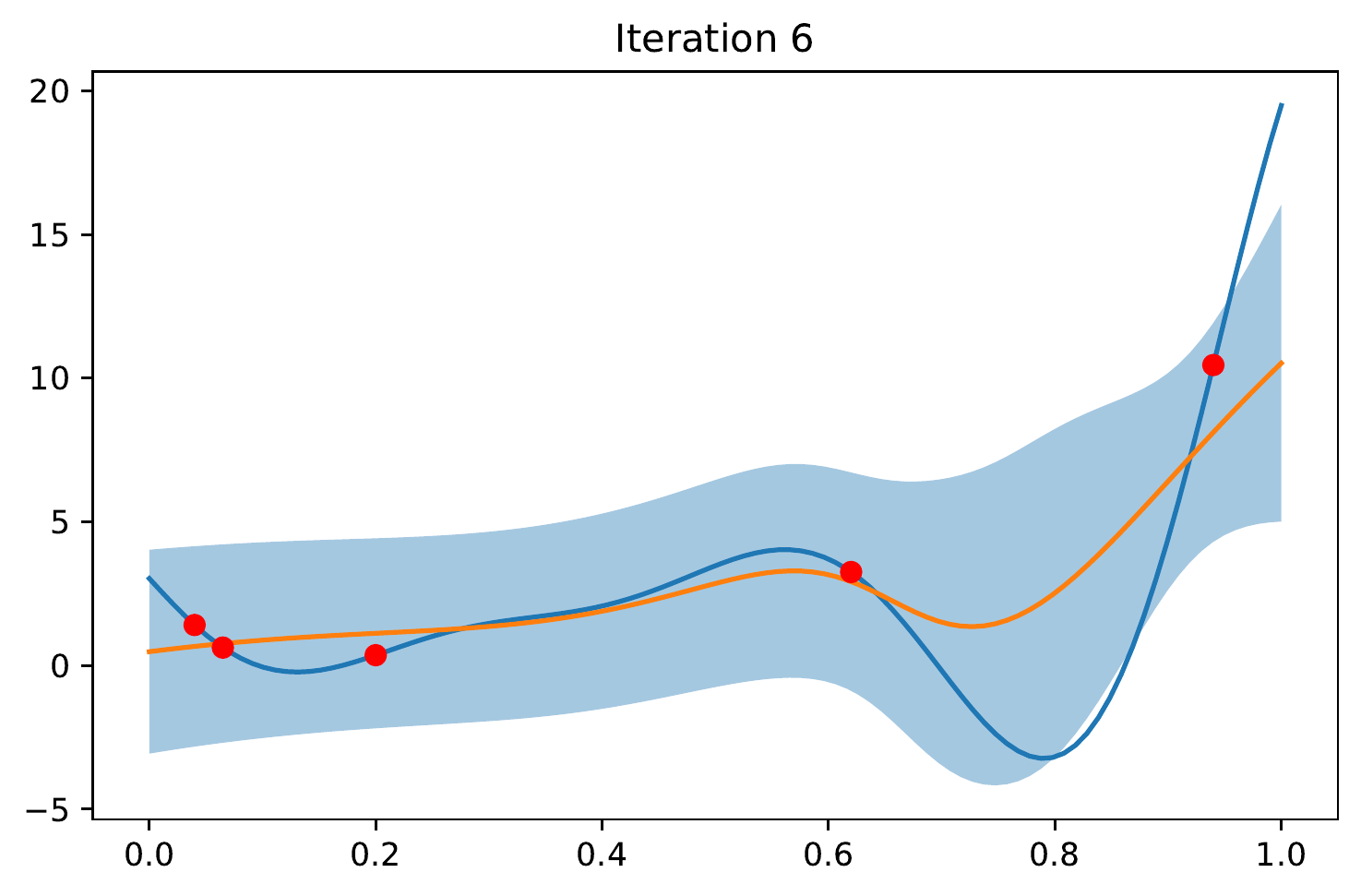}
  \includegraphics[width=0.24\textwidth]{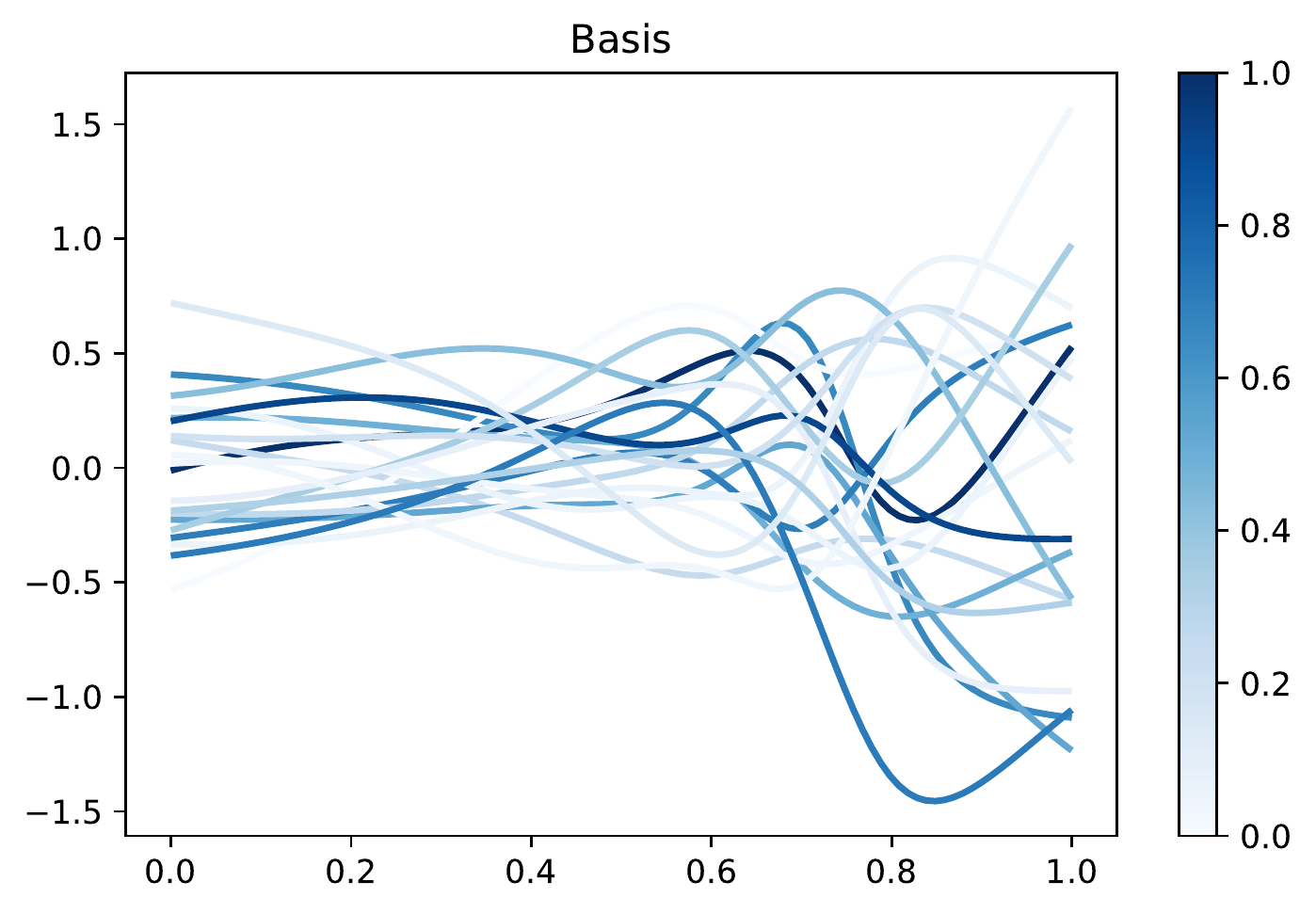}}
\hfill
  \subfigure[\texttt{ABLR L-BFGS fixed}]{
  \includegraphics[width=0.24\textwidth]{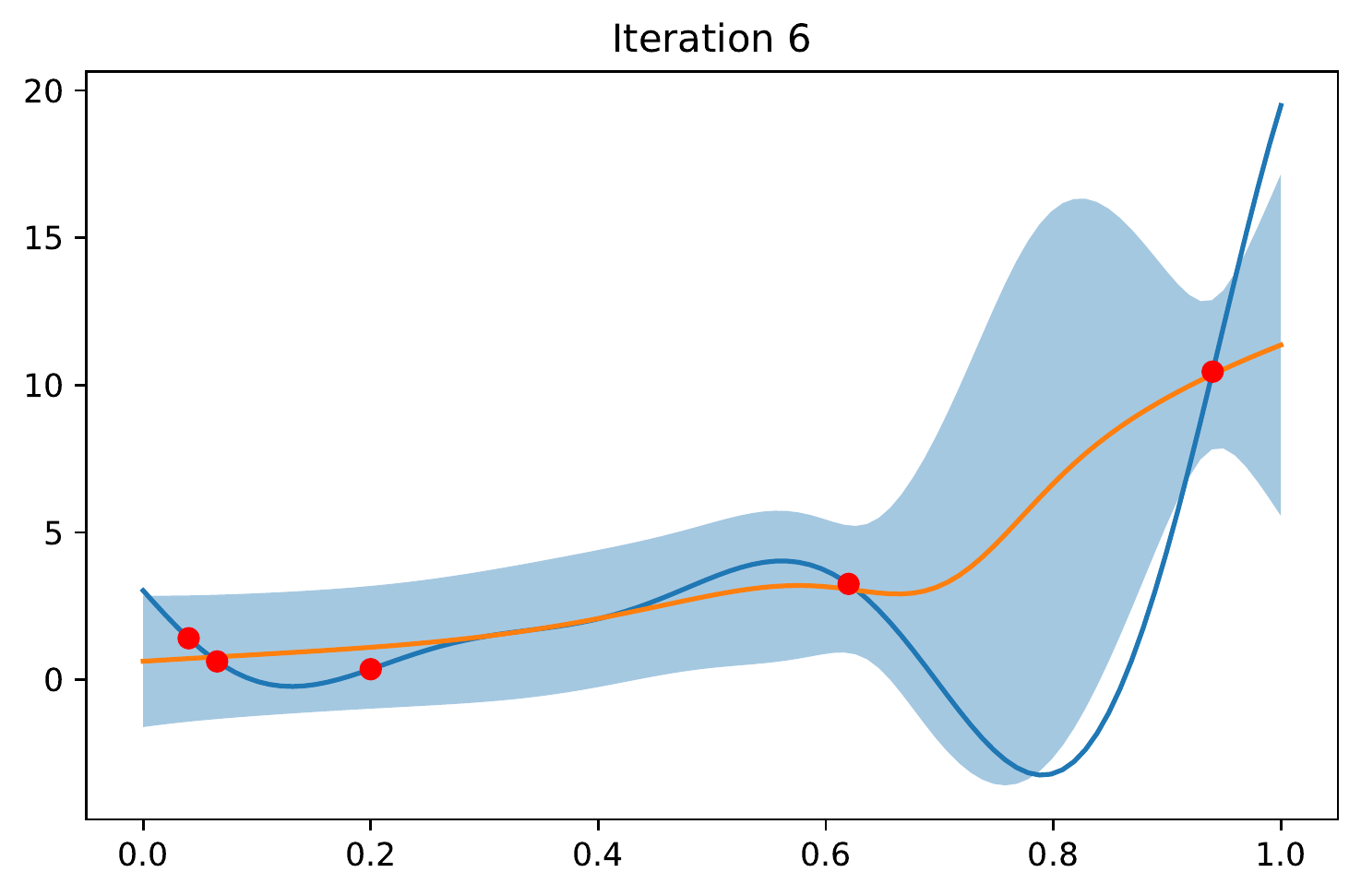}
  \includegraphics[width=0.24\textwidth]{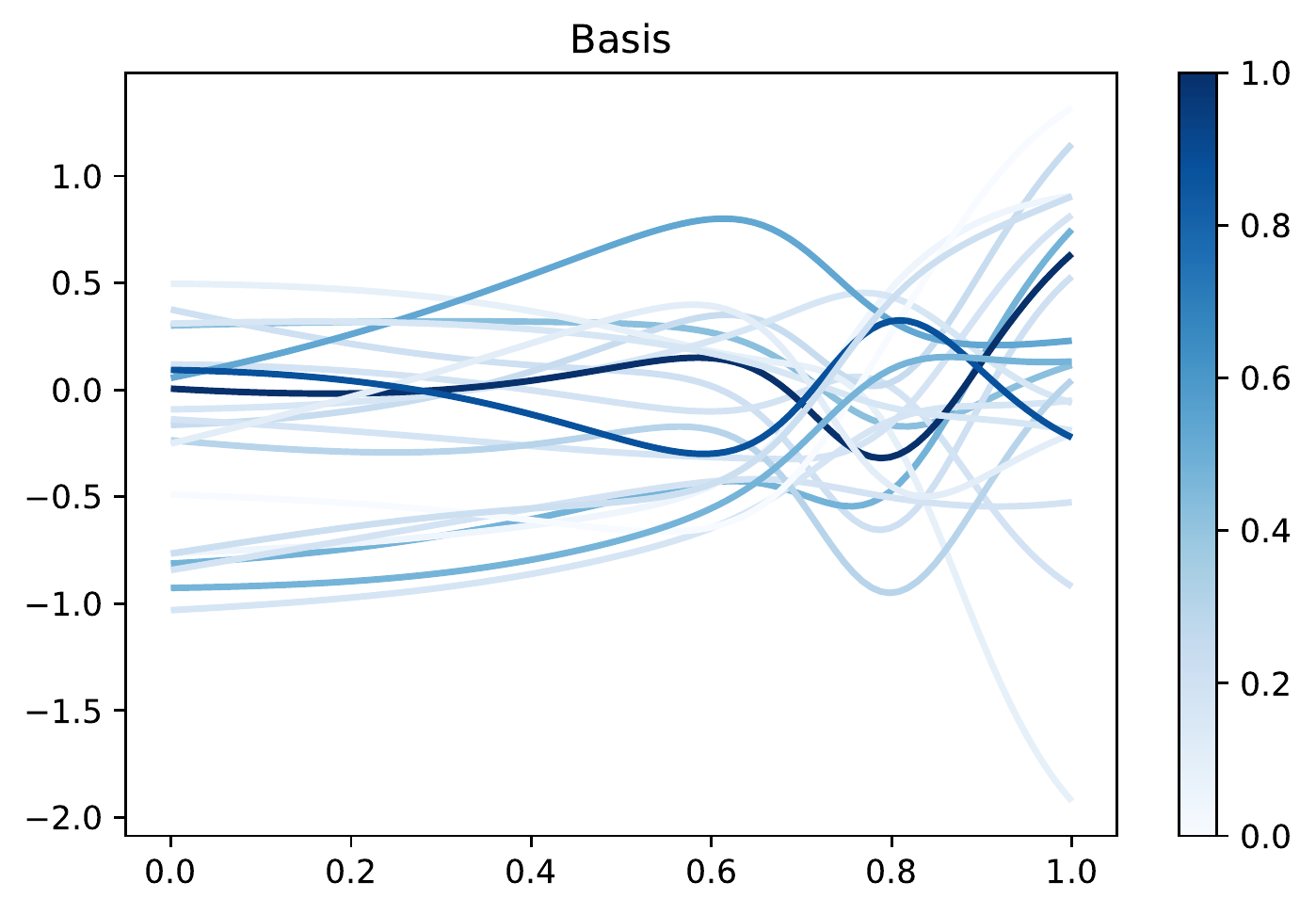}}
\hfill
\caption{ Comparison of the surrogate objective along with its basis functions after 5 total evaluations obtained by 4 different methods. In the first and the third panel the target function is shown in blue, the evaluations are the red points (which includes the same $5$ random points for all methods), and the surrogate predictive mean is shown in orange with two standard deviations in transparent light blue. The second and fourth panel show the basis functions used to fit the surrogate model for given algorithm with colour representing its relative weight.
}
  \label{fig:1d_unc_app}
\end{figure*}

\section{Bayesian Linear Regression: extra material for Section~\ref{sec:online_tr}}

For the details omitted in Section~\ref{sec:online_tr}, we continue with classical process of fitting Bayesian Linear regression (BLR) using Empirical Bayes, where one fits this probabilistic model by firstly integrating out weights $\wn$, which leads to the multivariate Gaussian model $\N(\mu(x_\n), \sigma^2(x_\n))$ for the next input $x_\n$ of the new task, whose mean and variance can be computed analytically~\citep{bishop2006pattern} 
\begin{eqnarray*}
 \mu(x_\n) &=& \bn f_\n^\top K^{-1}_\n (\Fn)^\top \yn,  \\
 \sigma^2(x_\n) &=& f_\n^\top K^{-1}_\n f_\n + \frac{1}{\bn},
\end{eqnarray*}
where $f_\n = \phi_{\z, b\downarrow}(x_\n)$ and $K_\n = \bn (\Fn)^\top \Fn + \An$. We use this as the input to the acquisition function, which decides the next point to sample. Before this step, we first need to obtain the parameters.  This model defines a Gaussian process, for which we can compute the covariance matrix in the closed-form solution. 

\begin{align*}
    \cov(y_i, y_j) &= \E\left[(\phi_{\z, b\downarrow}(x_i)^\top w + \epsilon_i)  (\phi_{\z, b\downarrow}(x_j)^\top w + \epsilon_j)^\top\right] \\
    &= \phi_{\z, b\downarrow}(x_i)^\top \E\left[ ww^\top \right] \phi_{\z, b\downarrow}(x_j) + \E\left[ \epsilon_i \epsilon_j\right] \\
    &= \phi_{\z, b\downarrow}(x_i)^\top \Diag(\mathbf{\alpha})^{-1} \phi_{\z, b\downarrow}(x_j) + \E\left[ \epsilon_i \epsilon_j\right],
\end{align*}
thus the covariance matrix has the following form $\Sigma_\n = \Fn \An^{-1} (\Fn)^\top + \bn^{-1}I_{N_\n}$. We obtain $\{\an\}_{i=1}^r$ and $\bn$ by minimizing the negative log likelihood of our model has, up to constant factors in the following form
\begin{equation*}
    \min_{\{\an\}_{i=1}^r,\bn} \frac{1}{2}\log |\Sigma_\n| + \yn^\top \Sigma_\n^{-1}\yn,
\end{equation*}
for which we use L-BGFS algorithm as this method is parameter-free and does not require to tune step size, which is a desired property as this procedure is run in every step of BO.  In order to speed up optimization and obtain linear scaling in terms of evaluation, one needs to include extra modifications, starting with decomposition of the matrix $(\Fn)^\top (\Fn) + \nicefrac{1}{\bn}\An$ as $L_\n L^{\top}_{\n}$ using Cholesky. This decomposition exists since $(\Fn)^\top (\Fn) + \nicefrac{1}{\bn}\An$  is always positive definite, due to semi-positive definiteness of  $(\Fn)^\top (\Fn)$ and positive diagonal matrix  $\nicefrac{1}{\bn}\An$. Final step is to use Weinstein–Aronszajn identity, and Woodbury matrix inversion identity, which implies that our objective can be rewritten to the the equivalent form 
\begin{align*}
  &\min_{\{\an\}_{i=1}^r,\bn} -\frac{(N_\n - r)}{2}\log (\bn) -  \frac{1}{2}\sum_{i=1}^r\log (\an) \\
  &+\sum_{i=1}^r \log ([L]_{ii}) + \frac{\bn}{2} (\|\yn \|^2 - \|L^{-1}_\n (\Fn)^\top \yn \|^2),
\end{align*}
which leads to overall complexity $\mathcal{O}(d^2\max\{N_\n, d\})$, which is linear in number of evaluations of the function $f_\n$.

\end{document}